\documentclass[letterpaper,journal]{IEEEtran}
\usepackage{amsmath,amsfonts}
\usepackage{algorithmic}
\usepackage{algorithm}
\usepackage{array}
\usepackage[caption=false,font=normalsize,labelfont=sf,textfont=sf]{subfig}
\usepackage{textcomp}
\usepackage{stfloats}
\usepackage{url}
\usepackage{verbatim}
\usepackage{graphicx}
\usepackage{booktabs}
\usepackage{cite}
\usepackage{hyperref}
\usepackage{multirow}
\begin{document}

\title{Forecasting Anomaly Precursors \\ with Uncertainty-Aware Time-series Ensembles}

\author{Hyeongwon~Kang,
        Jinwoo~Park,
        Seunghun~Han,~\IEEEmembership{Student Member,~IEEE,}
        and~Pilsung~Kang,~\IEEEmembership{Member,~IEEE}%
\thanks{Hyeongwon Kang is with the Department of Industrial \& Management Engineering, Korea University, 126-16 Anam-dong 5-ga, Seongbuk-gu, Seoul, Republic of Korea (e-mail: hyeongwon\_kang@korea.ac.kr).}%
\thanks{Jinwoo Park and Pilsung Kang are with the Department of Industrial Engineering, Seoul National University, Gwanak-ro 1, Gwanak-gu, Seoul, Republic of Korea (e-mail: \{jinwoo\_park, pilsung\_kang\}@snu.ac.kr).}%
\thanks{Seunghun Han is with LG CNS, 71, Magokjungang 8-ro, Gangseo-gu, Seoul, Republic of Korea (e-mail: seunghun.han@lgcns.com).}%
}

\markboth{Manuscript submitted to IEEE Transactions on Neural Networks and Learning Systems}%
{Kang \MakeLowercase{\textit{et al.}}: Forecasting Anomaly Precursors with Uncertainty-Aware Time-series Ensembles}


\maketitle

\begin{abstract}

Detecting anomalies in time-series data is critical in domains such as industrial operations, finance, and cybersecurity, where early identification of abnormal patterns is essential for ensuring system reliability and enabling preventive maintenance. However, most existing methods are reactive—they detect anomalies only after they occur—and lack the capability to proactively signal early warning signs.
In this paper, we propose FATE (Forecasting Anomalies with Time-series forecast Ensembles), a novel unsupervised framework for detecting Precursors-of-Anomaly (PoA) by quantifying predictive uncertainty from a diverse ensemble of time-series forecasting models. Unlike prior approaches that rely on reconstruction errors or require ground-truth labels, FATE anticipates future values and leverages ensemble disagreement to signal early signs of potential anomalies—without access to target values at inference time.
To rigorously evaluate PoA detection, we introduce Precursor Time-series Aware Precision and Recall (PTaPR), a new metric that extends the traditional Time-series Aware Precision and Recall (TaPR) by jointly assessing segment-level accuracy, within-segment coverage, and temporal promptness of early predictions. This enables a more holistic assessment of early warning capabilities that existing metrics overlook.
Experiments on five real-world benchmark datasets show that FATE achieves an average improvement of +19.9\%p in PTaPR AUC and +20.02\%p in early detection F1-score, outperforming baselines while requiring no anomaly labels. These results demonstrate the effectiveness and practicality of FATE for real-time, unsupervised early warning in complex time-series environments.

\end{abstract}


\section{Introduction}
\label{sec:sec1}

\begin{figure}[!t]
    \centering  
    \includegraphics[width=\columnwidth]{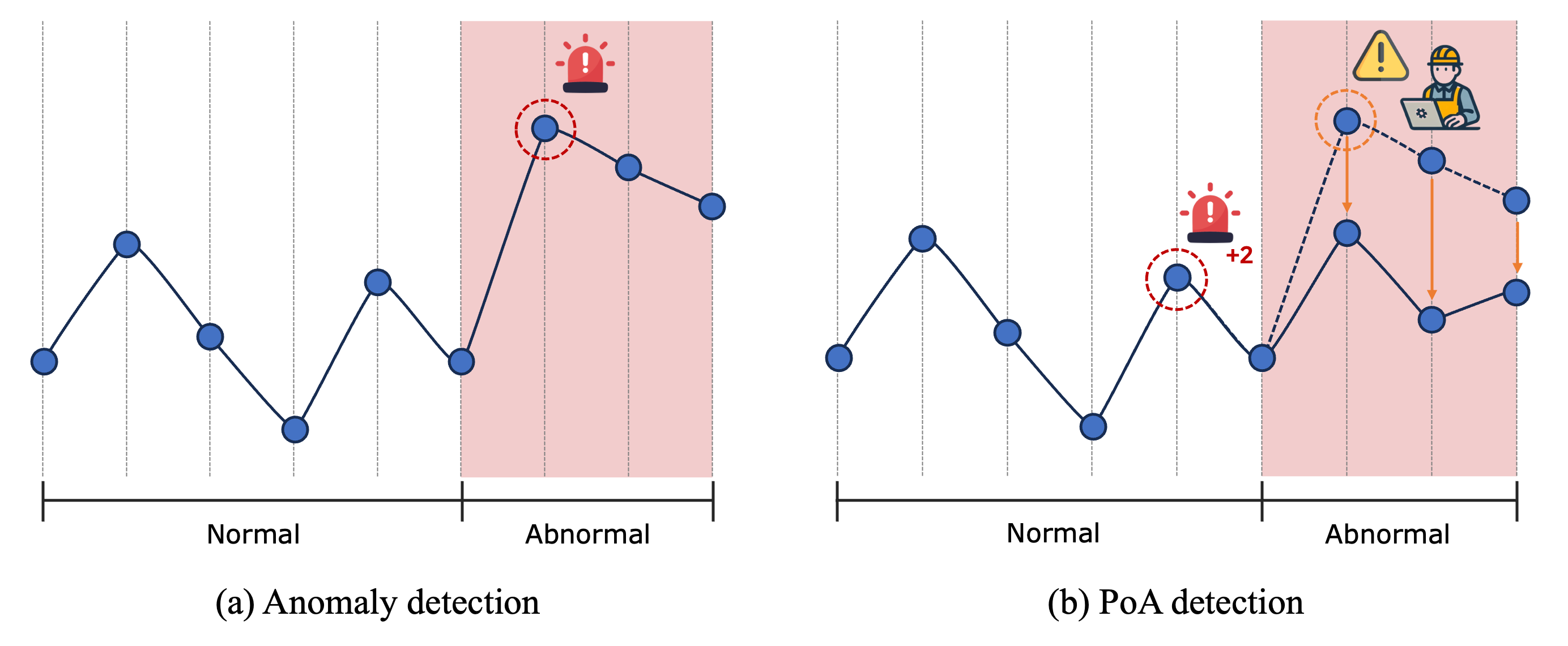}
    \caption{Comparison between traditional anomaly detection and Precursor-of-Anomaly (PoA) detection. (a) Traditional methods detect anomalies after their occurrence, limiting the ability to prevent failures. (b) PoA detection identifies early warning signals prior to anomaly onset, enabling proactive intervention.}
    \label{fig:PoA_detection}
\end{figure}

\IEEEPARstart{T}{ime-series anomaly detection} refers to the task of identifying abnormal data points that deviate from expected patterns in sequential data collected over time. This typically involves analyzing historical observations in time-series to determine whether a particular time point falls outside the normal range.
Recent advances in data storage and processing infrastructure have made the collection and analysis of time-series data increasingly accessible across a wide range of industrial domains. In particular, the rapid advancement in industrial automation and sensor technologies has led to the continuous generation of large volumes of time-series data across various domains such as manufacturing, logistics, transportation, and energy.
Accurate detection and monitoring of anomalies in such data are considered critical for preventing potential accidents and minimizing economic losses \cite{RN1, RN2, RN3}. As a result, research on effective time-series anomaly detection methods has gained significant attention, and its importance continues to grow \cite{RN4, RN5, RN6}.

Much of the existing research is concentrated on the analysis of data related to anomalies that have already happened, with the goal of identifying abnormal situations in the current or previous datasets \cite{RN16, RN17}. This method, commonly known as current anomaly detection, depends on analyzing post-anomaly data, as shown in \autoref{fig:PoA_detection}(a). While such methods are advantageous in enabling rapid detection and response once an anomaly has occurred, they inherently lack the capability to anticipate and prevent anomalies beforehand \cite{RN7}.
For example, in manufacturing processes, if a machine fault is detected only after an anomaly has manifested, production downtime and related losses may become unavoidable. Therefore, there is a growing need for early detection and prevention of anomalies before they occur.

To address this limitation, there has been growing interest in Precursor-of-Anomaly (PoA) detection, which aims to predict anomalies before their actual occurrence by identifying early-warning patterns or signals in the input data \cite{RN7, RN49, RN50, RN51}. As illustrated in \autoref{fig:PoA_detection}(b), PoA detection focuses on recognizing current patterns that may indicate the emergence of future abnormal events. Unlike conventional methods, this paradigm enables preventive actions by detecting early signs prior to the onset of actual anomalies.
Typical applications include early detection of mechanical faults in manufacturing processes or anticipating abnormal conditions in power plant equipment. Since PoA detection must operate under uncertainty without access to future observations, ensuring reliable predictive performance remains a major challenge. Furthermore, many existing methods rely heavily on the accuracy of a single forecasting or reconstruction model, which may lead to unstable performance in complex real-world settings. Another critical limitation lies in the lack of explicit control over lead time—i.e., how far in advance the anomaly is detected—which is essential for enabling timely and actionable responses. Despite these challenges, PoA detection holds considerable potential for mitigating unexpected failures and minimizing economic losses through proactive intervention.

As noted earlier, most prior studies focus on identifying anomalies at or before the current time, and the corresponding evaluation metrics have also been designed to assess detection performance at those time points \cite{RN8, RN9}. In contrast, for PoA detection, key evaluation criteria include whether the anomaly was detected \textit{before} it occurred and how \textit{early} the detection was made. Early detection is crucial, as it enables timely interventions and the prevention of unexpected incidents.
However, existing metrics often fail to adequately capture the timeliness of detection or the importance of the detection time itself \cite{RN9}. Therefore, there is a need for new evaluation metrics that explicitly consider detection timing and promptness in order to more accurately assess the effectiveness of PoA detection. Such metrics can contribute to better evaluation and improvement of early warning systems in practice.

We propose a novel framework, FATE (Forecasting Anomalies with Time-series forecasting Ensemble), which detects precursors of anomalies by leveraging the outputs of ensemble forecasting models. To complement this, we introduce a new evaluation metric, PTaPR (Precursor Time-series Aware Precision and Recall), which explicitly addresses the limitations of existing metrics by accounting for detection timing and promptness. Unlike conventional anomaly detection methods, the proposed framework leverages ensemble results from multiple forecasting models to identify precursor signals. It assesses the likelihood of a precursor by measuring the \textit{uncertainty} arising from discrepancies among the predictions of different models. Here, uncertainty is defined as the variance of predicted values at each time point; the higher the variance, the greater the likelihood that a precursor is present.
The key intuition is that when inputs include early symptoms or latent anomalies, the forecasting models exhibit increased prediction variance due to the difficulty in accurately forecasting under such conditions. Based on this assumption, FATE provides an effective approach for detecting precursors by quantifying uncertainty from ensemble predictions.

The proposed evaluation metric, PTaPR (Precursor Time-series Aware Precision and Recall), extends the existing TaPR (Time-series Aware Precision and Recall) by incorporating additional scoring strategies tailored for precursor detection. PTaPR integrates three scoring schemes to comprehensively assess the performance of detecting precursors-of anomalies.
This metric not only evaluates whether anomalies were detected but also assesses how accurately and how early precursors were identified before the actual anomaly occurred. By explicitly reflecting the timeliness and the specific detection point, PTaPR offers a more precise evaluation of the practical effectiveness of precursor detection systems.

To validate the effectiveness of the proposed FATE framework, we conducted a comparative evaluation using the PTaPR metric across five widely used benchmark datasets in time-series anomaly detection. The performance of FATE was compared against eight unsupervised baseline models. Despite relying solely on uncertainty scores without access to future target values, the proposed model outperformed conventional baseline models that utilize actual future observations.
Specifically, FATE achieved improvements in PTaPR AUC of 20.39\%p on the PSM dataset, 33.21\%p on MSL, 35.66\%p on SMAP, and 6.46\%p on SMD. Additionally, we conducted a detailed analysis to separate the effects of early precursor detection and downstream reward mechanisms, confirming that the model also excels in pure anomaly detection performance.
These results demonstrate that FATE surpasses baseline methods in both precursor prediction and anomaly detection, highlighting its practical potential for real-time anomaly detection and forecasting systems.
The contributions of this study are summarized as follows:

\begin{itemize}
  \item We propose FATE (Forecasting Anomalies with Time-series forecast Ensembles), a novel ensemble-based anomaly detection framework that forecasts anomaly precursors by quantifying uncertainty from multiple prediction models.
  
  \item To evaluate precursor detection performance, we introduce PTaPR (Precursor Time-series Aware Precision and Recall), a new metric that extends TaPR by incorporating detection timeliness and precursor coverage.
  
  \item FATE achieves significant performance gains without using future target values, demonstrating its effectiveness for real-time anomaly prediction in industrial systems.
\end{itemize}

The remainder of this paper is organized as follows. In \autoref{sec:sec2}, we introduce the task and methodologies of time-series anomaly detection and forecasting, along with the TaPR metric, which forms the basis of the proposed PTaPR evaluation metric. \autoref{sec:sec3} provides a detailed explanation of the proposed FATE framework, and \autoref{sec:sec4} presents the PTaPR metric. \autoref{sec:sec5} describes the benchmark datasets, baseline models, and experimental settings. The experimental results, their interpretation, and detailed analysis are all presented in \autoref{sec:sec6}. Finally, \autoref{sec:sec7} concludes the paper with a summary and discussion of the contributions and implications of this study.

\section{Related Works}
\label{sec:sec2}
\subsection{Time-series Anomaly Detection}
\label{sec:sec2.1}

Time-series anomaly detection aims to identify abnormal patterns that deviate from normal temporal behavior and has been widely studied in industrial and scientific domains. Due to the ambiguity of anomaly definitions and the scarcity of labeled data, unsupervised approaches have become the dominant paradigm, learning normal patterns from unlabeled data and detecting anomalies via significant deviations.

Existing unsupervised methods can be broadly categorized into density-based, clustering-based, prediction-based, and reconstruction-based approaches. Density- and clustering-based methods identify anomalies as low-density or out-of-cluster samples \cite{RN11,RN12,RN13,RN14}. Prediction-based methods detect anomalies through large forecasting errors \cite{RN15,RN16}, while reconstruction-based methods interpret high reconstruction errors as anomalous behavior \cite{RN17,RN18,RN19}. Recently, Transformer-based models such as Anomaly Transformer and Variable Temporal Transformer have demonstrated strong performance by effectively capturing long-range temporal dependencies \cite{RN20,RN21}.

\subsection{Time-series Precursor-of-Anomaly Detection}
\label{sec:sec2.2}

% While most existing studies focus on detecting anomalies after their onset, relatively few have explored \textit{precursor-of-anomaly (PoA)} detection—predicting anomalies before they occur. However, for ensuring the reliability of safety-critical systems, early forecasting of anomalous behavior is essential for enabling preventive interventions.
% One of the earliest efforts in this direction is PAD (Precursor-of-Anomaly Detection) \cite{RN7}, which introduces a multitask learning framework based on Neural Controlled Differential Equations (NCDEs). PAD jointly learns current anomaly detection and binary precursor classification—determining whether an anomaly will occur within a fixed time window.
% In contrast, our approach goes beyond binary classification by estimating the likelihood of anomaly occurrence at specific future time points. This enables more fine-grained and practical PoA detection, improving both temporal resolution and real-world applicability.

While most existing studies focus on detecting anomalies after their onset, recent research has begun to explore \textit{precursor-of-anomaly (PoA)} detection, which aims to anticipate anomalies before they occur. This line of work is particularly important for safety-critical systems, where early warnings enable timely and preventive interventions.
Early approaches to PoA detection include PAD \cite{RN7}, which formulates the task as a multitask learning problem that jointly performs anomaly detection and binary precursor classification—predicting whether an anomaly will occur within a predefined future window. More recent studies have further investigated anomaly forecasting and early-warning mechanisms under various modeling assumptions. However, many existing methods still rely on coarse, window-level binary decisions or require access to future target values, limiting their temporal resolution and practical applicability.
In contrast, our approach estimates the likelihood of anomaly occurrence at specific future time points based solely on predictive uncertainty, without access to future observations. This enables fine-grained, time-specific PoA detection and provides a more practical and flexible early-warning framework for real-world time-series systems.

\subsection{Time-series Forecasting}
\label{sec:sec2.3}

Time-series forecasting aims to predict future values from historical temporal data and plays a critical role in domains such as finance, climate science, and production systems. While early approaches relied on statistical and classical machine learning models, recent advances in deep learning have substantially improved forecasting accuracy by capturing complex nonlinear and temporal dependencies.

In particular, Transformer-based models have emerged as state-of-the-art methods for time-series forecasting due to their ability to model long-range dependencies via self-attention mechanisms \cite{RN22,RN23,RN24,RN25}. Numerous variants have been proposed to enhance efficiency and robustness, addressing challenges such as long-term prediction and non-stationarity \cite{RN26,RN27,RN28,RN29,RN32,RN33}.

The strong predictive capability of these advanced forecasting models provides a natural foundation for proactive anomaly and precursor detection. Building on this insight, we leverage forecasting model ensembles to quantify predictive uncertainty for PoA detection.

\subsection{Time-series Aware Precision and Recall(TaPR)}
\label{sec:sec2.4}

Time-series Aware Precision and Recall (TaPR) \cite{RN10} is a segment-level evaluation metric designed for time-series anomaly detection, addressing the limitations of traditional point-wise precision and recall. Conventional metrics often fail to reflect the temporal continuity of anomalies and are sensitive to slight misalignments between prediction and ground truth.

TaPR evaluates model performance over labeled anomaly intervals by incorporating two scoring strategies. The first, \textit{detection scoring}, considers whether each ground-truth anomaly segment has been detected, treating partial matches as correct. The second, \textit{portion scoring}, quantifies how well the predicted anomaly region overlaps with the ground truth. This dual approach provides a more balanced assessment of detection quality in sequential contexts.

To account for temporal uncertainty, TaPR also introduces a \textit{temporal tolerance} component, allowing small timing errors in predictions to be accepted within a predefined margin. The final TaP and TaR scores are computed as weighted sums of the detection and portion scores, as shown in \autoref{eqn:eqn1}, where $\alpha$ controls the trade-off.

\begin{align}
\label{eqn:eqn1}
    \text{TaR} &= \alpha \times \text{TaR}^d + (1 - \alpha) \times \text{TaR}^p, \notag \\
    \text{TaP} &= \alpha \times \text{TaP}^d + (1 - \alpha) \times \text{TaP}^p.
\end{align}

Overall, TaPR offers a flexible and robust framework for evaluating anomaly detection in time-series, capturing both segment-wise correctness and temporal alignment.

\section{Proposed Method - FATE}
\label{sec:sec3}

We denote the time-series data as $T = \{x_1, \cdots, x_T\}$, where each $x_t \in \mathbb{R}^c$ represents a $c$-dimensional observation vector at time step $t$. This sequential data structure captures multiple variables evolving over time, forming a multivariate time-series.
The objective of PoA detection in this context is to estimate the likelihood of an anomaly occurring at a future time point, based solely on past and present patterns—without access to future observations. To achieve this, we adopt a sliding window approach, constructing input sequences of fixed length $L_x$ defined as $X_t = \{x_1^t, x_2^t, \cdots, x_{L_x}^t\}$, where $x_i^t \in \mathbb{R}^{c}$. Here, $X_t$ denotes the sequence of observations from time step $t - L_x + 1$ to $t$, allowing the model to consider a contiguous segment of past data as input.
Based on the input sequence $X_t$, various forecasting models generate an output sequence $Y_t = \{y_1^t, y_2^t, \cdots, y_{L_y}^t\}$, where $y_i^t \in \mathbb{R}^{c}$ corresponds to the predicted values for the next $L_y$ time steps. This allows the model to estimate the likelihood of future anomalies over the prediction horizon $\{t+1, t+2, \cdots, t+L_y\}$.

\subsection{Overall Architecture}
\label{sec:sec3.1}

\begin{figure*}[!t]
    \centering  
    \includegraphics[width=\textwidth]{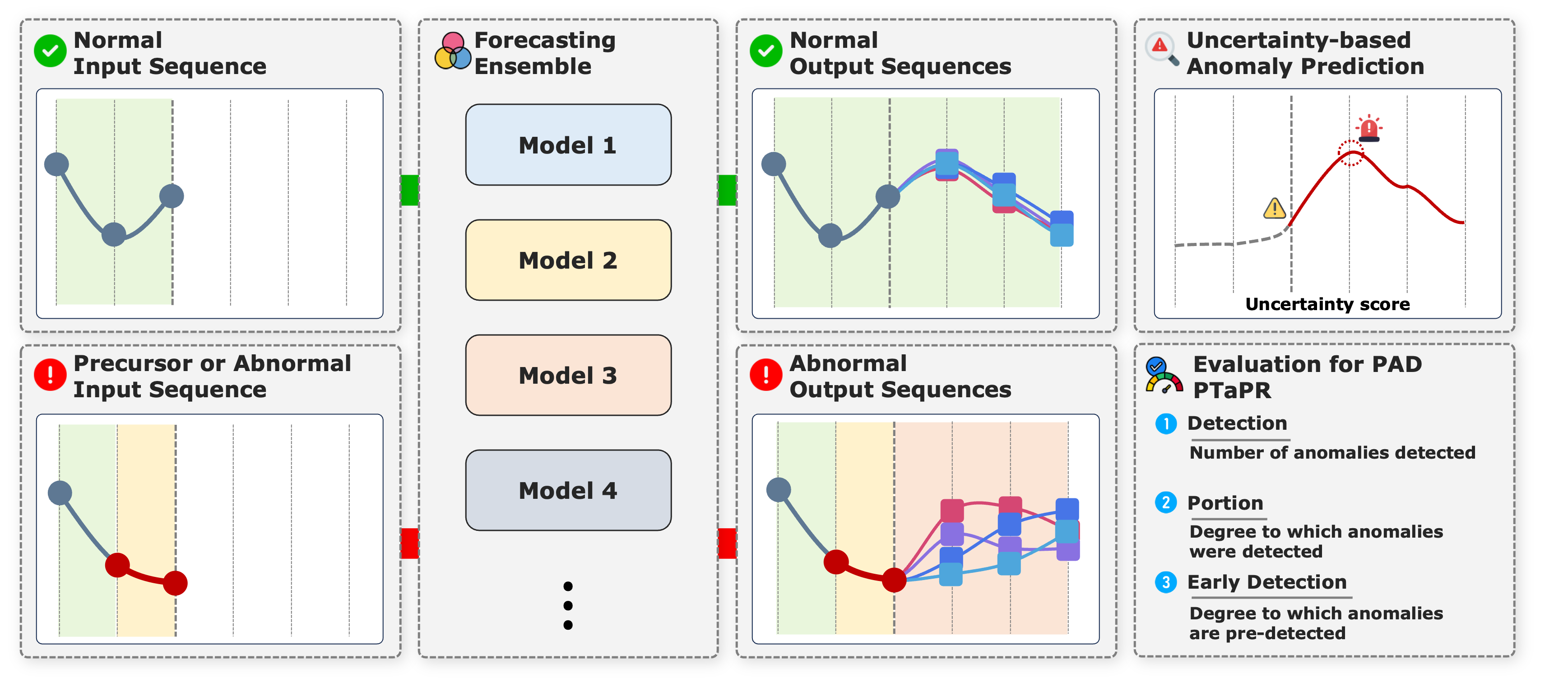}
    \caption{Overview of the proposed FATE framework. The input sequence is fed into an ensemble of forecasting models to predict future time steps. Uncertainty-based precursor detection is performed using the ensemble outputs, and early detection performance is quantitatively evaluated using the PTaPR metric.}
    \label{fig:FATE}
\end{figure*}

We propose a novel framework, FATE (Forecasting Anomalies with Time-series forecast Ensemble), that leverages prediction uncertainty from an ensemble of forecasting models to detect precursors of anomalies. \autoref{fig:FATE} illustrates the overall architecture.
FATE integrates multiple state-of-the-art forecasting models and computes the variance among their outputs at each prediction step to quantify uncertainty. This uncertainty score serves as an indicator of potential precursors, under the assumption that higher disagreement among models correlates with an increased likelihood of impending anomalies.
By identifying time steps with elevated uncertainty, FATE enables early warning predictions before actual anomalies occur. The ensemble-based approach enhances robustness and stability in estimating precursor likelihood across diverse time-series dynamics.
This framework is particularly suitable for real-time applications such as predictive maintenance, where timely and proactive responses are essential.

\subsection{Time-series Forecasting Models}
\label{sec:sec3.2}

Forecasting is the task of predicting a future output sequence $Y_t = \{y_1^t, y_2^t, \cdots, y_{L_y}^t\}, \; y_i^t \in \mathbb{R}^{c}$ based on a given input sequence $X_t = \{x_1^t, x_2^t, \cdots, x_{L_x}^t\}, \; x_i^t \in \mathbb{R}^{c}$. To derive uncertainty scores during this process, multiple forecasting algorithms are trained to optimize the mapping from the input sequence $X_t$ to the corresponding target sequence $Y_t$. Each model learns temporal patterns from historical data to enhance future prediction accuracy.

To ensure both diversity and robustness, we adopt a heterogeneous set of forecasting models as ensemble components. Since different algorithms are capable of capturing different data patterns and characteristics, they can complement each other by compensating for individual weaknesses \cite{RN39}. For example, one algorithm may be particularly effective in modeling certain trends or seasonality, while another may perform better in capturing irregular or abrupt changes. Moreover, heterogeneous models tend to exhibit resilience to different types of errors, making an ensemble of multiple models more robust than any individual model \cite{RN40}.
This ensemble approach provides more stable and reliable forecasting results than relying on a single model. We select the top-$K$ forecasting algorithms with the best performance to construct the ensemble. These top-performing models are considered to have high predictive accuracy and strong generalization capabilities across diverse forecasting scenarios.

Although top-$K$ models are used in this study, the choice of ensemble components can be adapted based on the dataset characteristics and the specific forecasting task.

\subsection{Uncertainty score for PoA}
\label{sec:sec3.3}

The uncertainty score used for PoA detection is computed based on the variance among the outputs of multiple forecasting models. This score quantitatively captures the degree of disagreement among predictions from different models at a specific time point. A higher level of disagreement—i.e., greater variance—implies a higher uncertainty score, which may indicate an increased likelihood of an impending anomaly.

The uncertainty score $U_i^t$ at time step $t$ for variable $i$ is defined as the variance of the predictions from $M$ forecasting models, as formally defined in \autoref{eqn:eqn2}:

\begin{gather}
\label{eqn:eqn2}
    \centering
        U_i^t = \frac{1}{M-1} \sum_{m=1}^M \left(y_{mi}^t - \bar{y}_i^t\right)^2, \quad \bar{y}_i^t = \frac{1}{M} \sum_{m=1}^M y_{mi}^t,
\end{gather}

Here, $y_{mi}^t$ denotes the prediction of model $m$ for variable $i$ at time $t$, and $\bar{y}_i^t$ is the mean prediction across all models. The uncertainty score reflects how much individual predictions deviate from the ensemble mean, thus indicating the consistency among model outputs. Greater disagreement results in a higher uncertainty score, which corresponds to lower confidence in the prediction at that time step.

Typically, as the prediction horizon increases, uncertainty tends to rise due to the inherent difficulty of long-term forecasting and reduced model reliability \cite{RN41, RN42}. Consequently, applying a fixed threshold uniformly across all prediction steps may lead to increased false positives at later time points. Conversely, using a high threshold might suppress the detection of meaningful precursor signals. 

To address this issue, the uncertainty scores are normalized to enable the consistent application of a unified threshold across all prediction steps. This normalization process compensates for the systematic increase in uncertainty over time and ensures score comparability across different horizons. Specifically, the mean $\mu_i$ and standard deviation $\sigma_i$ of the uncertainty scores at prediction step $i$ across all input windows $w$ are computed as shown in \autoref{eqn:eqn3}:

\begin{gather}
\label{eqn:eqn3}
    \centering
        \mu_i = \frac{1}{N} \sum_{t=1}^N U_i^t, \quad \sigma_i = \sqrt{\frac{1}{N} \sum_{t=1}^N \left(U_i^t - \mu_i\right)^2},
\end{gather}

Here, $N$ denotes the total number of input sequences used to compute the uncertainty scores at prediction step $i$—that is, the number of uncertainty values generated across different sliding windows. These values provide the statistical basis for estimating the uncertainty distribution at each prediction horizon. Based on this distribution, the normalized uncertainty score $U_{i\text{-norm}}^t$ is defined in \autoref{eqn:eqn4} as:

\begin{gather}
\label{eqn:eqn4}
    \centering
        U_{i-\text{norm}}^t = \frac{U_i^t - \mu_i}{\sigma_i},
\end{gather}

This normalization corrects for horizon-specific variations and allows for a consistent application of thresholds across all prediction steps, thereby improving the robustness and consistency of anomaly precursor detection.

By enabling fair comparison across time steps, normalized uncertainty scores enhance the reliability of PoA prediction. The normalized uncertainty score makes the uncertainty at each prediction point relatively comparable, which contributes to the reliability of anomaly prediction.

\subsection{PoA Detection}
\label{sec:sec3.4}

To enable reliable PoA detection, we normalize the uncertainty scores at each prediction step using the mean ($\mu_i$) and standard deviation ($\sigma_i$) computed from a held-out validation set. This prevents overfitting to training data and ensures a more objective evaluation on the test set by aligning uncertainty values to a consistent reference distribution. The normalized scores allow consistent thresholding across time steps and datasets.

Selecting an appropriate threshold is crucial for optimizing the model's anomaly detection performance. The threshold is determined using a best-F1 search strategy, which evaluates multiple candidate values on the test set and selects the one that achieves the optimal balance between precision and recall. The F1-score, which represents the harmonic mean of precision and recall, serves as a comprehensive metric for evaluating detection performance. In our case, the F1-score is computed based on the PTaPR evaluation metric.

\begin{figure}[!t]
    \centering  
    \includegraphics[width=\columnwidth]{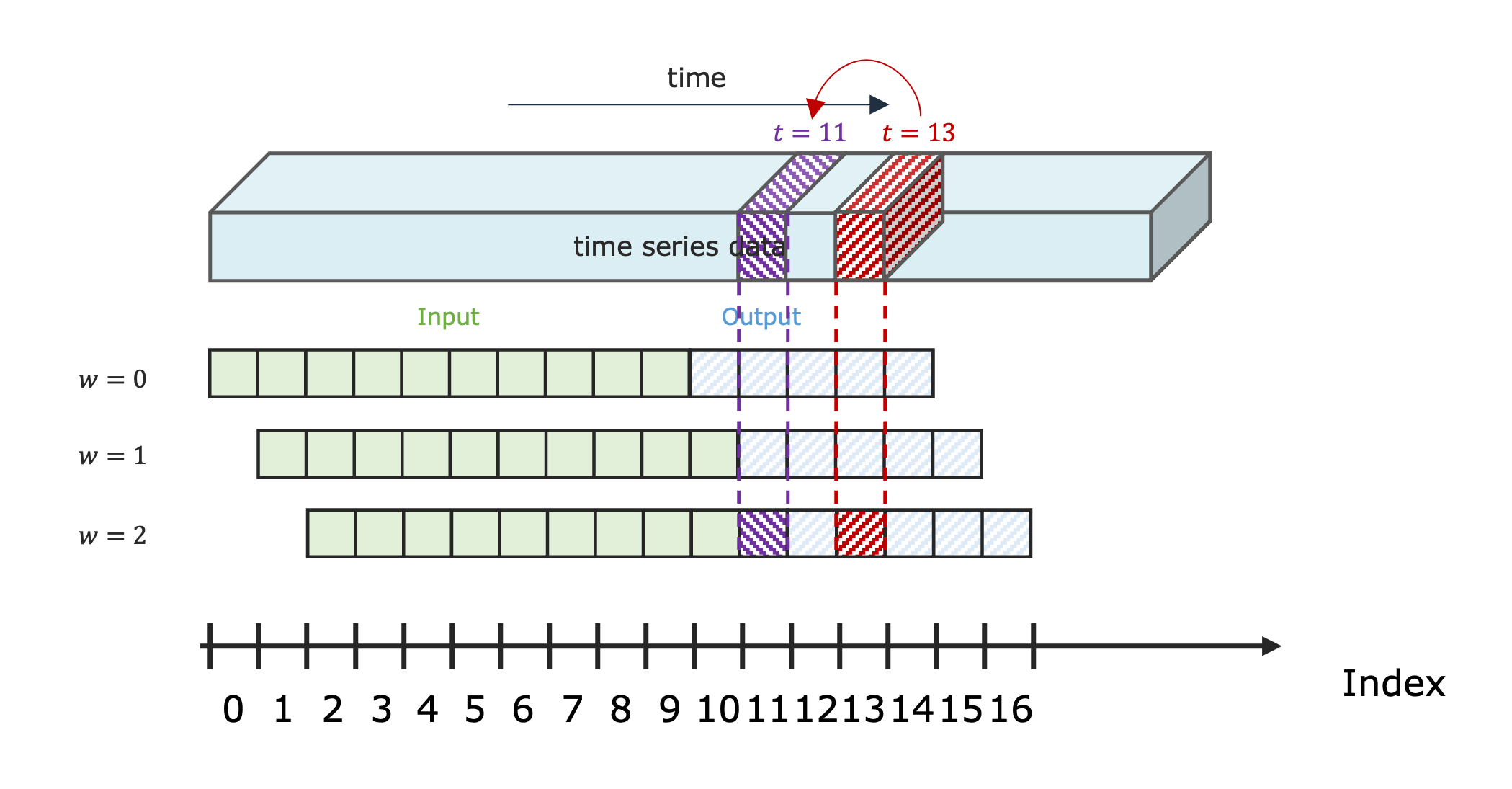}
    \caption{Example of predicting anomaly precursors at future time steps based on input at time $t$. The third window ($w=2$) forecasts $L_y=5$ future steps from current time $t+L_x-1=11$, and a precursor is detected at time step 13.}
    \label{fig:Labeling}
\end{figure}

Finally, the index $t$ of the input sequence can be used to track which future time steps are associated with the predictions made at the current time $t + L_x - 1$. For example, \autoref{fig:Labeling} illustrates a case where the input sequence length is 10 and the prediction horizon is 5. Assuming that an anomaly precursor is predicted in the third window ($w=2$), the input corresponds to time steps from $t = 2$ to $t + L_x - 1 = 11$. Based on this input, the model forecasts the future from time step 12 to 16 and detects a precursor at time step 13. This implies that the model, at the current time step 11, successfully predicts a potential precursor of an anomaly that may occur at time step 13.

The proposed FATE framework extends beyond conventional anomaly detection by enabling proactive identification of anomaly precursors. While ensemble-based uncertainty estimation is a well-established concept, its application to real-time PoA detection remains largely unexplored. FATE bridges this gap by leveraging a predictive-performance-based ensemble of forecasting models to estimate uncertainty and detect precursors without access to future labels, offering robust and lead-time-controllable early warning capabilities.

\section{Proposed evaluation metric – PTaPR}
\label{sec:sec4}

\begin{figure}[!t]
    \centering  
    \includegraphics[width=\columnwidth]{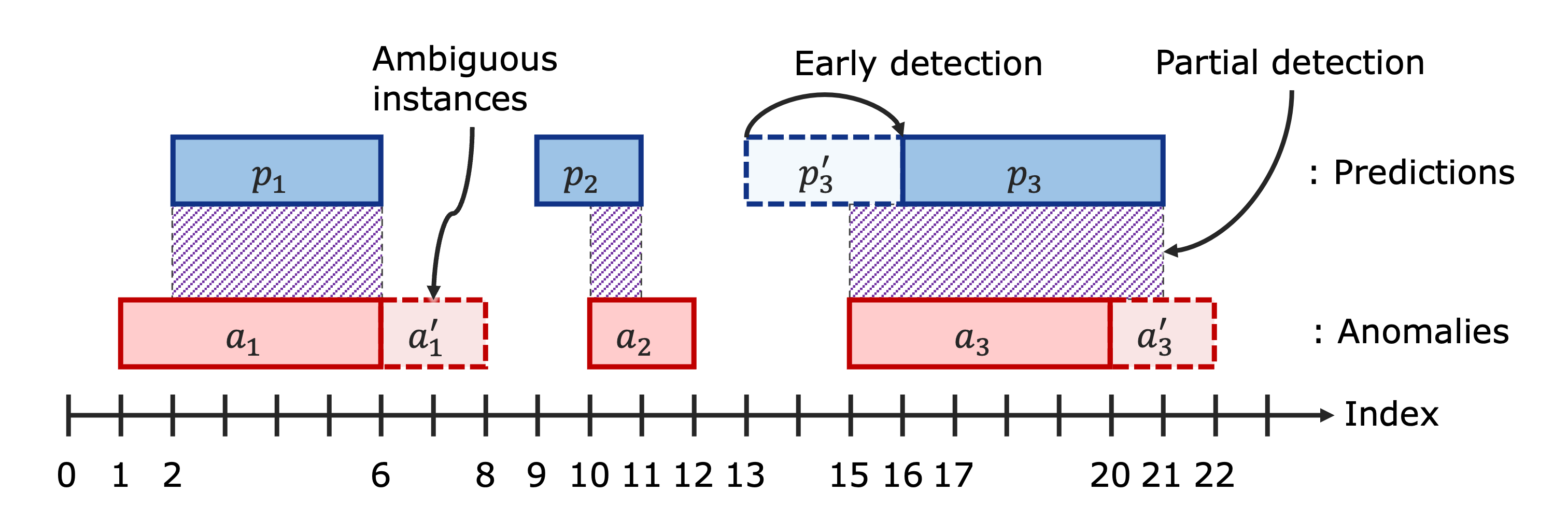}
    \caption{Illustration of early detection, partial detection, and ambiguous instances in the PTaPR metric. Anomalies and predictions are represented as instance-level segments along the x-axis. For example, anomaly $a_1$ spans indices 1 to 6, while early prediction $p'_3$ detects an anomaly between indices 13 and 16.}
    \label{fig:PTaPR}
\end{figure}

We propose a novel evaluation metric, PTaPR (Precursor Time-series Aware Precision and Recall), tailored to assess PoA detection in time-series data. PTaPR incorporates \textit{early detection}, \textit{partial detection}, and \textit{ambiguous instances}—i.e., those persisting beyond labeled anomaly segments—to reflect the temporal characteristics of time-series anomalies. As shown in \autoref{fig:PTaPR}, PTaPR captures the interaction between predicted segments ($p$) and ground-truth anomalies ($a$); for instance, $p_3$ precedes $a_3$ (early detection), while $p_1$ overlaps $a_1$ (partial detection). Unlike existing TaPR, PTaPR explicitly accounts for detection timing and segment continuity, offering a more robust and actionable evaluation of time-series anomaly detection models.

\subsection{Terminology}
\label{sec:sec4.1}

An anomaly segment is a sequence of consecutive instances $a = \{t, t+1, \cdots, t+l-1\}$, where $t$ is the start index and $l$ the segment length. The full set is denoted by $A = \{a_1, a_2, \cdots, a_n\}$.
A prediction segment $p = \{t', t'+1, \cdots, t'+l'-1\}$ represents anomalous instances detected by the model, with $P = \{p_1, p_2, \cdots, p_m\}$ as the set of all predictions.
The precursor segment $p' = \{t'', t''+1, \cdots, t'-1\}$ refers to the interval preceding $p$ where an anomaly precursor is detected. The collection is denoted as $P' = \{p'_1, p'_2, \cdots, p'_m\}$.
Finally, $a' = \{t+l, t+l+1, \cdots, t+l+\delta-1\}$ represents ambiguous instances following an anomaly, with $A' = \{a'_1, a'_2, \cdots, a'_n\}$ as the full set.

\subsection{Precursor Time-series Aware Recall}
\label{sec:sec4.2}

\begin{figure*}[t!]
    \centering  
    \includegraphics[width=\textwidth]{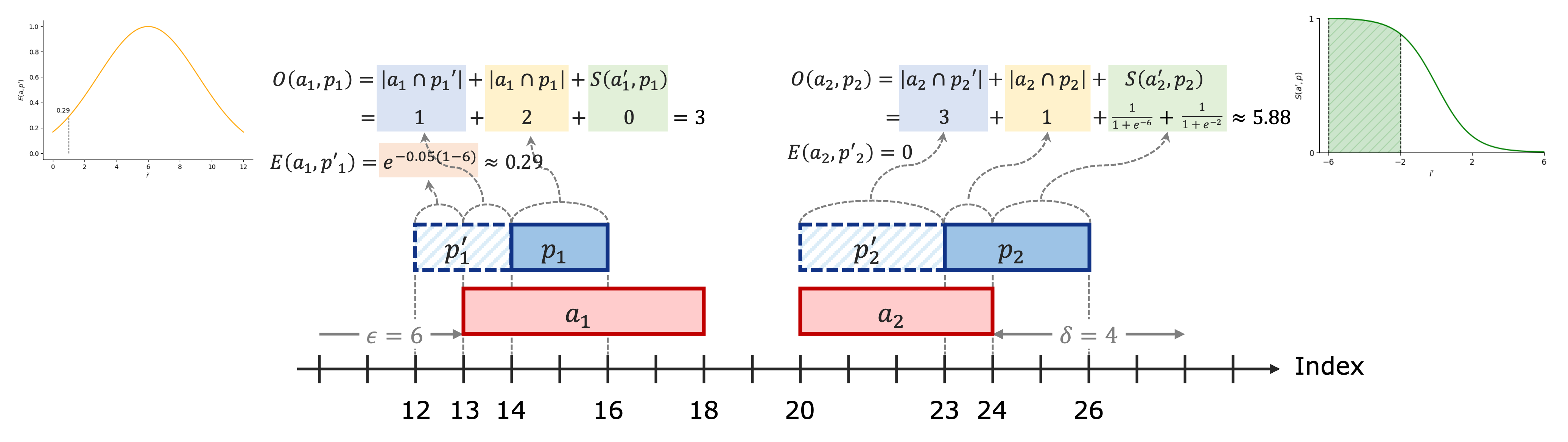}
    \caption{Example of PTaPR computation for anomaly segments ($a_1, a_2$) and prediction segments ($p_1, p_2$). The overlap score $O(a, p, p')$ is the sum of detected anomalies $|a \cap p|$, early predictions $|a \cap p'|$, and the ambiguous instance score $S(a', p)$. For $a_1$, the overlap score is 3 with an early detection reward of 0.29. For $a_2$, the overlap score is approximately 5.88, reflecting the contribution of the sigmoid-weighted ambiguous instances.}
    \label{fig:PTaPR_example}
\end{figure*}

Precursor Time-series Aware Recall (PTaR) is an extended recall metric designed to more comprehensively evaluate anomaly and precursor detection performance in time-series data. Unlike traditional recall, which only considers whether anomalies are detected, PTaR incorporates the quality of in-segment detection and the timeliness of early predictions, thereby better reflecting the sequential nature of time-series anomalies. As shown in \autoref{eqn:eqn5}, PTaR is formulated as a weighted sum of three components, each capturing a distinct aspect of precursor detection performance:

\begin{align}
\label{eqn:eqn5}
\text{PTaR} ={} & \alpha \cdot \text{PTaR}^d + \beta \cdot \text{PTaR}^p + \gamma \cdot \text{PTaR}^e, \notag \\
                & \quad \text{s.t. } \alpha + \beta + \gamma = 1,
\end{align}

Here, $\alpha$, $\beta$, and $\gamma$ are weighting factors for each component: $\text{PTaR}^d$ denotes segment-level detection accuracy, $\text{PTaR}^p$ represents within-segment detection coverage, and $\text{PTaR}^e$ captures early prediction quality.

1) Segment-level Detection Rate ($\text{PTaR}^d$):
This component measures the proportion of anomaly segments that are successfully detected, as defined in \autoref{eqn:eqn6}:

\begin{align}
\label{eqn:eqn6}
\text{PTaR}^d &= \frac{|A^d(\theta)|}{|A|}, \notag \\
A^d(\theta) &= \left\{ a \mid a \in A \text{ and } \frac{\sum_{p \in P} O(a, p, p')}{|a|} \geq \theta \right\},
\end{align}

where $\theta \in [0,1]$ is a user-defined threshold that specifies the minimum required overlap ratio for a segment $a$ to be considered successfully detected. A higher $\theta$ imposes a stricter criterion, requiring a greater proportion of the anomaly segment to be covered by the predicted or precursor segments.

The overlap score $O(a, p, p')$ quantifies the degree to which the predicted segment $p$ and its corresponding precursor segment $p'$ cover the anomaly segment $a$, including ambiguous instances. It is formally defined in \autoref{eqn:eqn7} as:

\begin{gather}
\label{eqn:eqn7}
    \centering
      O(a, p, p') = |a \cap p'| + |a \cap p| + S(a', p),
\end{gather}

The overlap score includes early predictions ($p'$), in-segment detections ($p$), and ambiguous instances ($a'$) following the anomaly. The ambiguous instance score $S(a', p)$ reflects the extent to which the model continues to detect abnormal behavior beyond the labeled anomaly segment. As defined in \autoref{eqn:eqn8}, the score is computed using a sigmoid-based weighting scheme:

\begin{gather}
\label{eqn:eqn8}
    \centering
      S(a', p) = \sum_{i \in (a' \cap p)} \frac{1}{1 + e^{i'}}, \quad i' = -6 + \frac{12(i - t_{a'})}{\delta - 1},
\end{gather}

where $i'$ is the normalized distance from the end of the abnormal segment to the ambiguous instance $i$, which is scaled to the range $[-6, 6]$ and then applied with a sigmoid function. This imposes a penalty on predictions that are far from the original abnormal segment.

2) Within-segment Detection Coverage ($\text{PTaR}^p$):
The second component, $\text{PTaR}^p$, quantifies the within-segment detection coverage by measuring how extensively each anomaly segment is covered by the predicted and precursor segments. To prevent overestimation caused by excessive overlap, the detection ratio is capped at 1 using a $\min$ function. Formally, it is defined in \eqref{eqn:eqn9}:

\begin{gather}
\label{eqn:eqn9}
    \centering
      \text{PTaR}^p = \frac{1}{|A|} \sum_{a \in A} \min\left(1, \frac{\sum_{p \in P} O(a, p, p')}{|a|}\right),
\end{gather}

This formulation ensures that the metric faithfully reflects the consistency of anomaly segment coverage without bias from redundant detections.

3) Early Prediction Reward ($\text{PTaR}^e$):
The third component, $\text{PTaR}^e$, quantifies the model’s ability to anticipate anomalies by assigning higher scores to predictions made at an appropriate temporal distance from the anomaly onset. It is formulated as the average early detection reward over all anomaly segments, as shown in \autoref{eqn:eqn10}:

\begin{align}
\label{eqn:eqn10}
\text{PTaR}^e &= \frac{1}{|A|} \sum_{a \in A} E(a, p'), \notag \\
E(a, p') &= e^{-k(i' - \epsilon)^2}, \quad i' = t_a - i, \quad \text{where } i < t_a,
\end{align}

Here, i' is the time gap between a precursor prediction and the start of the anomaly, $\epsilon$ denotes the optimal lead time, and $k$ controls the decay curvature. This formulation rewards timely predictions while penalizing those that are either too early or too late.

As illustrated in \autoref{fig:PTaPR_example}, the components are computed for each anomaly segment. Suppose $p_1$ overlaps with $a_1$ and $p'_1$ captures one early prediction; then $O(a_1, p_1, p'_1) = 3$ and $E(a_1, p'_1) \approx 0.29$. Similarly, $p_2$ overlaps with $a_2$ and captures some ambiguous instances, yielding $O(a_2, p_2, p'_2) \approx 5.88$ and $E(a_2, p'_2) = 0$. If both segments meet the overlap threshold $\theta$, we obtain $\text{PTaR}^d = 1$, $\text{PTaR}^p = (0.6 + 1)/2 = 0.8$, and $\text{PTaR}^e = (0.29 + 0)/2 = 0.145$. With equal weights ($\alpha = \beta = \gamma = \frac{1}{3}$), the final PTaR is calculated as $\text{PTaR} \approx \frac{1}{3}(1 + 0.8 + 0.145) = 0.65$.

This demonstrates that PTaR not only evaluates whether an anomaly is detected, but also incorporates the promptness of precursor prediction and consistency of in-segment coverage. As such, PTaR provides a more comprehensive and temporally aware evaluation than traditional recall metrics in the context of time-series anomaly detection.

\subsection{Precursor Time-series Aware Precision}
\label{sec:sec4.3}

Precursor Time-series Aware Precision (PTaP) extends traditional precision by evaluating the alignment between predicted and actual anomaly segments, while also accounting for internal consistency and the reliability of early predictions. It is defined in \autoref{eqn:eqn11}:

\begin{align}
\label{eqn:eqn11}
\text{PTaP} &= \alpha \cdot \text{PTaP}^d + \beta \cdot \text{PTaP}^p + \gamma \cdot \text{PTaP}^e, \notag \\
           & \quad \text{s.t. } \alpha + \beta + \gamma = 1,
\end{align}

1) Segment-level Prediction Accuracy ($\text{PTaP}^d$): 
This measures the proportion of predicted segments that sufficiently overlap with ground-truth anomalies. As shown in \autoref{eqn:eqn12}, $\text{PTaP}^d$ is defined as the ratio of such valid predictions to the total number of predicted segments:

\begin{align}
\label{eqn:eqn12}
\text{PTaP}^d &= \frac{|P^c(\theta)|}{|P|}, \notag \\
P^c(\theta) &= \left\{p \mid p \in P \text{ and } \frac{\sum_{a \in A} O(a, p, p')}{|p|} \geq \theta \right\}
\end{align}

Here, $\theta$ denotes the minimum required overlap ratio, and $O(a, p, p')$ is the overlap score defined previously.

2) In-segment Precision ($\text{PTaP}^p$):
This measures the average proportion of correctly predicted instances within each prediction segment. As shown in \autoref{eqn:eqn13}, it is defined as the mean overlap ratio between each prediction segment and the corresponding anomaly segments:
% The second component, in-segment precision ($\text{PTaP}^p$), measures the average proportion of correctly predicted instances within each prediction segment. As shown in \autoref{eqn:eqn13}, it is defined as the mean overlap ratio between each prediction segment and the corresponding anomaly segments, capped at 1 to prevent overestimation from excessive overlap:

\begin{gather}
\label{eqn:eqn13}
    \centering
      \text{PTaP}^p = \frac{1}{|P|} \sum_{p \in P} \min\left(1, \frac{\sum_{a \in A} O(a, p, p')}{|p|}\right),
\end{gather}

The $\min$ operator ensures that the value does not exceed 1, preventing overestimation from excessive overlap.

3) Early Prediction Reliability ($\text{PTaP}^e$):
This component quantifies the model's ability to anticipate anomalies in advance. As defined in \autoref{eqn:eqn14}, it computes the average early detection reward across all precursor segments:
% The third component, early prediction reliability ($\text{PTaP}^e$), quantifies the model’s ability to anticipate anomalies before they occur. As defined in \autoref{eqn:eqn14}, it computes the average early detection reward across all precursor segments:

\begin{gather}
\label{eqn:eqn14}
    \centering
      \text{PTaP}^e = \frac{1}{|P'|} \sum_{p' \in P'} E(a, p'),
\end{gather}

Here, $E(a, p')$ denotes the reward score for early detection, and $P'$ is the set of precursor segments preceding each prediction segment in $P$.

To illustrate the computation of PTaP, consider the example shown in \autoref{fig:PTaPR_example}. For $p_1$ and $p_2$, the overlap scores with ground-truth anomalies are $O(a_1, p_1, p'_1) = 3$ and $O(a_2, p_2, p'_2) \approx 5.88$, respectively. Given segment lengths of 2 and 3, the normalized overlap ratios exceed $\theta$ for both, resulting in $\text{PTaP}^d = 1$. Next, both segments yield capped internal precision scores of 1.0, giving $\text{PTaP}^p = 1$. For early prediction reliability, only $p'_1$ yields a reward of approximately 0.29, while $p'_2$ contributes nothing, leading to $\text{PTaP}^e = 0.145$. The final PTaP score is calculated as: $\text{PTaP} = \frac{1}{3}(1 + 1 + 0.145) \approx 0.72$.

The outcome illustrates the comprehensive nature of PTaP in evaluating both detection accuracy and temporal responsiveness.

\subsection{PTaPR Final Score}
\label{sec:sec4.4}

PTaPR (Precursor Time-series Aware Precision and Recall) is defined as the harmonic mean of PTaR (Recall) and PTaP (Precision), offering a balanced evaluation of the model’s ability to accurately identify anomaly segments and their precursors. As shown in \autoref{eqn:eqn15}, the metric is computed as:

\begin{gather}
\label{eqn:eqn15}
    \centering
      \text{PTaPR} = \frac{2 \times \text{PTaR} \times \text{PTaP}}{\text{PTaR} + \text{PTaP}},
\end{gather}

This formulation provides a comprehensive measure of the model's performance in detecting both anomaly segments and their precursors, quantifying its overall detection capability.

When computing PTaPR, a minimum overlap threshold $\theta$ is applied to determine whether a predicted segment is considered a valid detection. The value of $\theta$ varies from 0 to 1, and the corresponding F1-scores are calculated at each threshold to assess the robustness of the model across different detection criteria. This approach extends the idea of $PA\%K$ \cite{RN8}, where a prediction is only considered correct if it covers a certain proportion of the anomaly segment.

To compute PTaPR, we introduce a tunable overlap threshold $\theta$, which determines whether a predicted segment is considered to correctly match a ground-truth anomaly. Specifically, a prediction is treated as a valid detection if the normalized overlap score exceeds $\theta$. By varying $\theta$ in the range $[0, 1]$, we generate a set of PTaPR values across multiple detection strictness levels. This approach generalizes the idea behind the $PA\%K$ metric \cite{RN8}, where partial detection is deemed valid only when a fixed percentage of an anomaly segment is correctly predicted. Instead of selecting a single $\theta$, we aggregate performance over the full range by computing the Area Under the Curve (AUC) of the PTaPR scores with respect to $\theta$. This AUC serves as our final evaluation criterion, summarizing the model’s consistency and robustness in precursor-of-anomaly detection across varying levels of tolerance. A higher AUC indicates that the model performs reliably across both lenient and strict evaluation thresholds, and thus, exhibits stable detection capabilities even under shifting requirements for overlap quality.

\section{Experimental Setting}
\label{sec:sec5}

\subsection{Datasets}
\label{sec:sec5.1}

\begin{table}[t!]
\centering
\caption{Details of Benchmarks}
\label{tab:Details of benchmarks}
\renewcommand{\arraystretch}{0.8}
\setlength\tabcolsep{5pt}
\begin{tabular}{lrrrrr}
\toprule
                         & \textbf{SWaT} & \textbf{PSM} & \textbf{SMD} & \textbf{MSL} & \textbf{SMAP} \\ \midrule
Variables                & 51      & 25      & 38      & 55     & 25      \\
Number of entities       & 1       & 1       & 28      & 27     & 54      \\
\#Train (0.7)            & 346{,}500 & 115{,}921 & 495{,}884 & 40{,}822 & 96{,}603 \\
\#Valid (0.3)            & 148{,}500  & 49{,}680  & 212{,}521 & 17{,}495 & 41{,}401  \\
\#Test (labeled)         & 449{,}919 & 87{,}841  & 708{,}420 & 73{,}729 & 427{,}617 \\
Anomaly (\%)             & 11.98   & 27.8    & 4.16    & 10.72  & 13.13   \\
\bottomrule
\end{tabular}
\end{table}

To evaluate the effectiveness of the proposed methodology, we conducted experiments using five widely adopted benchmark datasets for unsupervised anomaly detection: SWaT, PSM, MSL, SMAP, and SMD. These datasets are commonly used in the literature and consist of unlabeled training sets and labeled test sets containing injected anomalies for evaluation purposes. The labeled test sets are designed to include predefined anomaly segments to evaluate the anomaly detection performance of the model.

(1) SWaT (Secure Water Treatment) \cite{RN34} contains time-series data collected from 51 sensors over an 11-day period in a water treatment testbed. The last four days of data include a total of 36 attack scenarios that represent abnormal system behavior.
(2) PSM (Pooled Server Metrics) \cite{RN35} is a 25-dimensional dataset internally collected from multiple application server nodes at eBay. It includes aggregated metrics such as CPU usage, memory allocation, and network traffic, providing a comprehensive benchmark for testing model robustness in server environments.
(3) SMD (Server Machine Dataset) \cite{RN36} consists of 5 weeks of multivariate data with 38 metrics collected from 28 server machines in a large-scale internet company.
(4) MSL (Mars Science Laboratory rover) and (5) SMAP (Soil Moisture Active Passive satellite) \cite{RN16} are datasets released by NASA, containing telemetry data with 55 and 25 dimensions, respectively. These datasets include anomaly labels extracted from Incident Surprise Anomaly (ISA) reports related to spacecraft monitoring systems.

Among the selected datasets, SMD, MSL, and SMAP consist of multiple sub-datasets, and models were trained and evaluated separately on each sub-dataset. A summary of the datasets, including the number of variables, number of observations, train/validation/test splits, and the proportion of anomalies in the test set, is presented in \autoref{tab:Details of benchmarks}.

% \begin{figure}[!t]
%     \centering  
%     \includegraphics[width=\columnwidth]{Fig6_Dataset_split.png}
%     \caption{The process of splitting the dataset. 70\% of the training dataset is used to train the model, and the remaining 30\% is utilized as the forecasting evaluation dataset to evaluate the forecasting performance against normal patterns. The test dataset contains the data on which the model should actually detect anomalies and is not used in the training process.}
%     \label{fig:Dataset}
% \end{figure}

We consider a forecasting model trained in an unsupervised setting, where no anomaly labels are available. The model learns patterns from normal data during training time and detects anomalies at inference time by identifying deviations from these patterns.
To evaluate forecasting performance fairly, we split the training data: 70\% is used for training, and 30\% is reserved as an anomaly-free forecasting evaluation set. This allows us to assess the model's ability to predict future values without contamination from anomalies.
The test set, containing real anomalies, is used solely for evaluation and remains unseen during training. It serves as a benchmark to examine whether the model, trained only on normal patterns, can identify anomaly precursors in unseen data. This setup ensures a reliable and realistic evaluation of both forecasting accuracy and anomaly detection effectiveness in unsupervised settings.

\subsection{Time-series Forecasting Models}
\label{sec:sec5.2}

We construct an ensemble using a diverse set of state-of-the-art time-series forecasting models, covering Transformer-based, convolutional, and linear architectures. These models are designed to capture complementary temporal characteristics such as long-range dependencies, seasonality, and non-stationary patterns \cite{RN22,RN23,RN25,RN26,RN27,RN28,RN29,RN30,RN31,RN32,RN33,RN37}.

All forecasting models were trained on the training dataset and evaluated on a separate forecasting validation set. Predictive performance was assessed using standard error metrics, including Mean Absolute Error (MAE) and Mean Squared Error (MSE), by comparing predicted outputs $\hat{Y}_t$ with the observed values $Y_t$.

\begin{table}[t]
\centering
\caption{Time-series forecasting performance (MSE and MAE) across five datasets. Best scores are in bold and underlined.}
\label{tab:forecasting_results}
\renewcommand{\arraystretch}{0.95}
\scriptsize
\setlength\tabcolsep{4pt}
\resizebox{\columnwidth}{!}{%
\begin{tabular}{lccccccc}
\toprule
\textbf{Method} & \textbf{Metric} & \textbf{PSM} & \textbf{SWaT} & \textbf{MSL} & \textbf{SMAP} & \textbf{SMD} & \textbf{Average} \\
\midrule
\multirow{2}{*}{Transformer} & MSE & 0.0080 & 728.0698 & 0.0146 & 0.0224 & 0.0084 & 145.6227 \\
                             & MAE & 0.0601 & 7.1352   & 0.0618 & 0.0719 & 0.0271 & 1.4712 \\
\midrule
\multirow{2}{*}{Informer}    & MSE & 0.0056 & 639.0894 & 0.0177 & 0.0235 & 0.0096 & 127.8273 \\
                             & MAE & 0.0449 & 6.6135   & 0.0680 & 0.0774 & 0.0356 & 1.3679 \\
\midrule
\multirow{2}{*}{Reformer}    & MSE & 0.0090 & 756.6053 & 0.0190 & 0.0222 & 0.0088 & 151.3318 \\
                             & MAE & 0.0657 & 7.3256   & 0.0694 & 0.0661 & 0.0275 & 1.5109 \\
\midrule
\multirow{2}{*}{Autoformer}  & MSE & 0.0011 & 29.2684  & 0.0356 & 0.0434 & 0.0100 & 5.8720 \\
                             & MAE & 0.0257 & 1.2082   & 0.0984 & 0.1302 & 0.0402 & 0.3005 \\
\midrule
\multirow{2}{*}{Crossformer} & MSE & 0.0004 & 88.6510  & 0.0017 & 0.0162 & 0.0086 & 17.7354 \\
                             & MAE & 0.0120 & 2.3140   & 0.0173 & 0.0345 & 0.0229 & 0.4801 \\
\midrule
\multirow{2}{*}{ETSformer}   & MSE & 0.0044 & 76.5729  & 0.0075 & 0.1043 & 0.0233 & 15.3419 \\
                             & MAE & 0.0480 & 1.7300   & 0.0531 & 0.2447 & 0.1002 & 0.4352 \\
\midrule
\multirow{2}{*}{FEDformer}   & MSE & 0.0008 & 0.5900   & 0.0540 & 0.0477 & 0.0172 & 0.1419 \\
                             & MAE & 0.0152 & 0.2289   & 0.1204 & 0.1387 & 0.0604 & 0.1127 \\
\midrule
\multirow{2}{*}{FiLM}        & MSE & 0.0004 & 0.2502   & 0.0014 & 0.0180 & 0.0080 & 0.0556 \\
                             & MAE & 0.0114 & 0.1088   & 0.0146 & 0.0404 & 0.0174 & 0.0385 \\
\midrule
\multirow{2}{*}{DLinear}     & MSE & 0.0004 & 0.2717   & 0.0014 & 0.0212 & 0.0130 & 0.0615 \\
                             & MAE & 0.0115 & 0.1136   & 0.0156 & 0.0702 & 0.0637 & 0.0549 \\
\midrule
\multirow{2}{*}{TimesNet}     & MSE & 0.0004 & \textbf{\underline{0.1402}} & 0.0021 & 0.0178 & 0.0079 & 0.0337 \\
                             & MAE & 0.0116 & \textbf{\underline{0.0885}} & 0.0177 & 0.0316 & 0.0137 & 0.0326 \\
\midrule
\multirow{2}{*}{Pyraformer}  & MSE & 0.0082 & 1.5806   & 0.0115 & 0.0200 & 3.2060 & 0.9653 \\
                             & MAE & 0.0541 & 0.4814   & 0.0535 & 0.0535 & 0.0548 & 0.1395 \\
\midrule
\multirow{2}{*}{PatchTST}    & MSE & \textbf{\underline{0.0004}} & 0.2224 & \textbf{\underline{0.0012}} & \textbf{\underline{0.0152}} & \textbf{\underline{0.0072}} & 0.0493 \\
                             & MAE & \textbf{\underline{0.0112}} & 0.1022 & \textbf{\underline{0.0128}} & \textbf{\underline{0.0283}} & \textbf{\underline{0.0127}} & 0.0334 \\
\bottomrule
\end{tabular}
}
\end{table}

As summarized in \autoref{tab:forecasting_results}, PatchTST consistently achieved the best overall performance with the lowest error rates across most datasets. FiLM and TimesNet also demonstrated strong forecasting accuracy, particularly on the SMD and SWaT datasets, respectively. Crossformer and DLinear followed closely, showing competitive results on datasets such as MSL and PSM. On the other hand, models like Reformer and ETSformer showed higher prediction errors, indicating relatively lower generalization.

Based on this evaluation, we selected the top-$K$ models ($K=5$) with the best forecasting performance to form our ensemble. The final ensemble was constructed by selecting high-performing models based on their average accuracy across datasets, while allowing for dataset-specific adjustments when a particular model demonstrated superior performance. By compensating for prediction errors that may arise in individual models on specific datasets, the ensemble contributes to improved overall accuracy. Combining various algorithms is a key strategy for ensuring model diversity and robustness, and performance gains can be expected from different combinations depending on the dataset and task objectives. In this study, we selected top-performing models with these considerations in mind, resulting in more stable and reliable forecasting outcomes.

\subsection{Implementation Details}
\label{sec:sec5.3}

In this study, the input sequence length was set to 100 and the prediction sequence length was set to 24. To ensure consistent segmentation over time, the window stride was set to 1. Each forecasting model was trained using the default hyperparameters provided in the original papers. For anomaly detection, the threshold was determined using the best F1 score search method. During model training and evaluation, a batch size of 32 was used. For PTaPR metric computation, the weights for each component—$\alpha$, $\beta$, and $\gamma$—were equally set to $1/3$. The optimal early detection window $\epsilon$ was selected via hyperparameter search tailored to the characteristics of each dataset. The sharpness parameter $k$ for the early detection reward curve was fixed at 0.001. All implementations were conducted using PyTorch \cite{RN38}, and training and evaluation were performed on a system equipped with four NVIDIA Tesla V100 GPUs (32GB each). These settings ensure the reproducibility and reliability of the experimental results, enabling a robust evaluation of the proposed methodology.

\subsection{Baselines}
\label{sec:sec5.4}

To evaluate our model, we compare it against several representative baselines in time-series anomaly detection, including LSTM-AE \cite{RN17}, LSTM-VAE \cite{RN18}, USAD \cite{RN19}, DAGMM \cite{RN12}, OmniAnomaly \cite{RN36}, Anomaly Transformer (AT) \cite{RN20}, and Variable Temporal Transformer (VTT) \cite{RN21}, each employing a different architectural paradigm. LSTM-AE employs an encoder-decoder structure to reconstruct normal sequences and detect anomalies based on reconstruction error. LSTM-VAE combines variational autoencoders with LSTM to model the probabilistic distribution of time-series data. DAGMM integrates an autoencoder with a Gaussian Mixture Model to detect anomalies using low-dimensional latent representations. OmniAnomaly leverages variational RNNs to model latent states in multivariate time series. Anomaly Transformer introduces a self-attention mechanism tailored for anomaly detection by learning relevance scores between time steps, enabling it to distinguish abnormal temporal dependencies from normal ones. VTT enhances anomaly detection by employing a dual attention architecture that concurrently captures dependencies across both time steps and variables.

VTT introduces two model architectures—VT-SAT and VT-PAT—which differ in their integration of temporal and variable attention: the former applies them sequentially, while the latter employs a parallel structure for joint attention.

While these models are widely used for anomaly detection, they are not specifically designed for forecasting anomalies before their occurrence. Among limited prior studies addressing this task, PAD (Precursor-of-Anomaly Detection) formulates PoA detection as a binary classification problem, predicting whether an anomaly will occur within a future time window. In contrast, our framework estimates the likelihood of anomaly occurrence at specific future time steps, making direct comparisons to PAD infeasible.

Given the absence of comparable PoA-focused models, we evaluate against standard anomaly detectors and apply our proposed PTaPR metric to assess both detection accuracy and early-warning capability. To ensure fairness, we credit baseline models with early detection when they generate continuous anomaly scores prior to the true anomaly onset. These instances are reflected in the early-detection component of PTaPR. This evaluation setup provides a more comprehensive and objective comparison, despite the baselines’ inherent limitations in proactive forecasting.

\section{Experimental Results}
\label{sec:sec6}
\subsection{Main Results}
\label{sec:sec6.1}

\begin{table*}[t]
\centering
\caption{Experimental results of the PTaPR metric on five time-series datasets. The best scores are in bold and underlined.}
\label{tab:PTaPR_results}
\renewcommand{\arraystretch}{0.4}
\setlength\tabcolsep{3pt}
\resizebox{\textwidth}{!}{%
\begin{tabular}{>{\centering\arraybackslash}p{1.2cm}|>{\centering\arraybackslash}p{1.2cm}|*{8}{>{\centering\arraybackslash}p{1.45cm}}|>{\centering\arraybackslash}p{1.45cm}}
\toprule
Dataset & Metric & \footnotesize{LSTM-AE} & \footnotesize{LSTM-VAE} & \footnotesize{USAD} & \footnotesize{DAGMM} & \footnotesize{OmniAnomaly} & \footnotesize{AT} & \footnotesize{VT-SAT} & \footnotesize{VT-PAT} & \footnotesize{FATE} \\
\midrule

\multirow{3}{*}{PSM}  
& $\text{F1}_0$  & 38.56  & 40.15  & 38.89  & 39.36  & 39.88  & 19.45  & 37.12  & 36.91  & \textbf{\underline{53.41}}  \\
& $\text{F1}_1$  & 16.67  & 16.52  & 16.19  & 15.86  & 15.93  & 4.09   & 14.91  & 15.72  & \textbf{\underline{34.31}}  \\
& AUC            & 28.59  & 28.64  & 28.34  & 28.52  & 28.74  & 7.90   & 25.93  & 26.66  & \textbf{\underline{50.64}}  \\
\midrule

\multirow{3}{*}{SWaT}  
& $\text{F1}_0$  & 29.64  & 30.79  & 32.62  & 29.83  & 32.36  & 15.06  & 26.61  & 26.76  & \textbf{\underline{36.29}}  \\
& $\text{F1}_1$  & 19.01  & 18.04  & 17.62  & 17.18  & 17.97  & 0.68   & 18.09  & 17.00  & \textbf{\underline{30.47}}  \\
& AUC            & 26.22  & 26.61  & 27.46  & 26.25  & 27.22  & 0.93   & 25.43  & 24.91  & \textbf{\underline{32.81}}  \\
\midrule

\multirow{3}{*}{MSL}  
& $\text{F1}_0$  & 22.53  & 22.13  & 22.14  & 21.83  & 21.93  & 13.51  & 23.44  & 22.11  & \textbf{\underline{48.65}}  \\
& $\text{F1}_1$  & 5.41   & 5.43   & 5.56   & 5.44   & 5.54   & 1.13   & 6.51   & 5.75   & \textbf{\underline{29.78}}  \\
& AUC            & 9.50   & 9.54   & 9.87   & 9.71   & 9.82   & 1.53   & 11.00  & 10.14  & \textbf{\underline{43.15}}  \\
\midrule

\multirow{3}{*}{SMAP}  
& $\text{F1}_0$  & 18.75  & 18.59  & 19.46  & 21.03  & 20.02  & 12.67  & 18.31  & 17.00  & \textbf{\underline{49.10}}  \\
& $\text{F1}_1$  & 5.28   & 5.31   & 5.42   & 6.29   & 5.50   & 1.22   & 6.09   & 4.97   & \textbf{\underline{33.45}}  \\
& AUC            & 7.95   & 7.97   & 8.08   & 9.87   & 9.00   & 1.80   & 8.91   & 7.73   & \textbf{\underline{44.16}}  \\
\midrule

\multirow{3}{*}{SMD}  
& $\text{F1}_0$  & 45.35  & 45.23  & 44.70  & 45.51  & 45.49  & 8.10   & 44.75  & 43.83  & \textbf{\underline{45.53}}  \\
& $\text{F1}_1$  & 23.86  & 23.74  & 23.49  & 23.58  & 23.69  & 2.24   & 24.59  & 24.00  & \textbf{\underline{31.11}}  \\
& AUC            & 35.38  & 35.26  & 34.95  & 35.25  & 35.31  & 3.70   & 36.10  & 35.16  & \textbf{\underline{41.80}}  \\
\bottomrule
\end{tabular}%
}
\end{table*}

We report the performance of FATE and several baseline models on five benchmark datasets in \autoref{tab:PTaPR_results}, using the PTaPR metric across three evaluation points: $\text{F1}_0$, $\text{F1}_1$, and AUC. Here, $\text{F1}_0$ corresponds to the F1-score when the minimum required overlap ratio $\theta$ is set to 0, which aligns with the point adjustment strategy and treats any detected point as valid. $\text{F1}_1$ denotes the F1-score with $\theta = 1$, equivalent to the conventional point-wise F1-score. The AUC is computed by varying $\theta$ from 0 to 1 and measuring the area under the resulting F1-score curve.

FATE operates in a fully unsupervised manner, relying solely on predictive uncertainty without access to future target values. Despite this constraint, it consistently outperforms all baseline methods across all five datasets. As shown in \autoref{tab:PTaPR_results}, FATE achieves substantial improvements in AUC, with gains of +22\%p on PSM, +5.35\%p on SWaT, +32.15\%p on MSL, +34.29\%p on SMAP, and +5.70\%p on SMD compared to the strongest competing baselines. These results demonstrate the strong generalization capability of FATE across diverse time-series characteristics. Notably, on the SWaT dataset—known for its complex cyber-physical anomaly patterns—several reconstruction-based baselines such as USAD and OmniAnomaly exhibit competitive performance. However, FATE still yields the highest AUC and $\text{F1}_1$ scores, underscoring its robustness in dense and overlapping anomaly settings. The Anomaly Transformer, a recent strong baseline in time-series anomaly detection, achieves relatively high $\text{F1}_0$ scores when $\theta = 0$, indicating sensitivity to any anomaly signals. However, its performance degrades sharply in both F1 and AUC as $\theta$ increases, revealing limitations in detecting temporally consistent or extended anomalies. In contrast, FATE maintains stable performance across varying overlap requirements, highlighting its advantage in detecting early signals of temporally extended anomalies.

While the proposed model demonstrated clear advantages in early detection as measured by the PTaPR metric, we acknowledge that these gains may be partially influenced by the reward mechanism for early detection. To investigate this more rigorously and ensure an objective performance comparison, \autoref{sec:sec6.2} presents additional evaluations using alternative metrics—namely, TaPR and PA\%K—which do not include the early detection reward component. These results show that FATE can effectively detect anomalies in complex datasets without relying on target values, demonstrating substantial practical competitiveness against strong baseline models.

\begin{table*}[t]
\centering
\caption{Experimental results for early detection on five time-series datasets. The best scores are in bold and underlined.}
\label{tab:early_detection}
\renewcommand{\arraystretch}{0.4}
\setlength\tabcolsep{3pt}
\resizebox{\textwidth}{!}{%
\begin{tabular}{>{\centering\arraybackslash}p{1.2cm}|>{\centering\arraybackslash}p{1.2cm}|*{8}{>{\centering\arraybackslash}p{1.5cm}}|>{\centering\arraybackslash}p{1.5cm}}
\toprule
Dataset & Metric & \footnotesize{LSTM-AE} & \footnotesize{LSTM-VAE} & \footnotesize{USAD} & \footnotesize{DAGMM} & \footnotesize{OmniAnomaly} & \footnotesize{AT} & \footnotesize{VT-SAT} & \footnotesize{VT-PAT} & \footnotesize{FATE} \\
\midrule

\multirow{3}{*}{PSM}  
& P   & 54.92  & 72.04  & 57.87  & 72.96  & \textbf{\underline{74.48}}  & 18.94  & 50.47  & 45.76  & 42.85  \\
& R   & 18.92  & 17.84  & 19.32  & 17.80  & 18.11  & 5.12   & 17.52  & 17.98  & \textbf{\underline{56.57}}  \\
& F1  & 28.15  & 28.59  & 28.96  & 28.61  & 29.14  & 8.06   & 26.01  & 25.81  & \textbf{\underline{48.76}}  \\
\midrule

\multirow{3}{*}{SWaT}  
& P   & 51.97  & 42.74  & 52.47  & 49.20  & 51.08  & 9.44   & 44.50  & \textbf{\underline{76.94}}  & 73.03  \\
& R   & 19.97  & 19.45  & 18.64  & 18.01  & 18.68  & 0.74   & 18.84  & 15.13  & \textbf{\underline{27.26}}  \\
& F1  & 28.85  & 26.74  & 27.51  & 26.37  & 27.36  & 1.37   & 26.47  & 25.29  & \textbf{\underline{39.66}}  \\
\midrule

\multirow{3}{*}{MSL}  
& P   & 14.83  & 14.61  & 14.49  & 14.26  & 14.62  & 9.12   & 15.88  & 14.76  & \textbf{\underline{41.13}}  \\
& R   & 11.52  & 11.86  & 12.61  & 12.76  & 12.85  & 1.36   & 13.78  & 12.79  & \textbf{\underline{56.50}}  \\
& F1  & 12.97  & 13.09  & 13.48  & 13.47  & 13.68  & 2.37   & 14.76  & 13.70  & \textbf{\underline{47.61}}  \\
\midrule

\multirow{3}{*}{SMAP}  
& P   & 13.28  & 13.29  & 13.70  & 16.32  & 15.50  & 9.00   & 14.54  & 13.27  & \textbf{\underline{38.21}}  \\
& R   & 13.36  & 12.86  & 13.31  & 14.67  & 13.89  & 1.91   & 12.15  & 10.80  & \textbf{\underline{64.64}}  \\
& F1  & 13.32  & 13.07  & 13.50  & 15.45  & 14.65  & 3.15   & 13.24  & 11.91  & \textbf{\underline{48.03}}  \\
\midrule

\multirow{3}{*}{SMD}  
& P   & 43.44  & 43.49  & 42.81  & 44.64  & \textbf{\underline{44.77}}  & 5.99   & 40.39  & 39.27  & 37.21  \\
& R   & 37.29  & 37.03  & 37.39  & 36.61  & 36.73  & 4.47   & 39.57  & 39.01  & \textbf{\underline{55.60}}  \\
& F1  & 40.13  & 40.00  & 39.92  & 40.23  & 40.35  & 5.12   & 39.98  & 39.14  & \textbf{\underline{44.58}}  \\
\bottomrule
\end{tabular}%
}
\end{table*}

\autoref{tab:early_detection} presents the results of early detection performance for both FATE and baseline methods. Specifically, we focus on the early-detection scoring components of the PTaPR metric—$\text{PTaP}^e$, $\text{PTaR}^e$, and their corresponding F1-score—to evaluate each model's ability to anticipate anomalies before they occur. While FATE showed relatively lower precision in early detection, it achieved notably high recall, successfully capturing most of the precursor signals. As a result, it outperformed the baseline models in terms of the overall F1-score, indicating superior early-warning capability. These findings underscore FATE's ability to anticipate anomalies, facilitating timely interventions that help prevent potential failures. In the context of real-time monitoring and early-warning systems, such performance suggests that the proposed framework could play a critical role in enhancing system reliability and preventing failures by enabling faster and more proactive responses.

\subsection{Performance Evaluation Without Early-detection Rewards}
\label{sec:sec6.2}

\begin{table*}[t]
\centering
\caption{Experimental results of the TaPR metric for anomaly detection on five time-series datasets. The best scores are in bold and underlined.}
\label{tab:TaPR_results}
\renewcommand{\arraystretch}{0.4}
\setlength\tabcolsep{3pt}
\resizebox{\textwidth}{!}{%
\begin{tabular}{>{\centering\arraybackslash}p{1.2cm}|>{\centering\arraybackslash}p{1.2cm}|*{8}{>{\centering\arraybackslash}p{1.5cm}}|>{\centering\arraybackslash}p{1.5cm}}
\toprule
Dataset & Metric & \footnotesize{LSTM-AE} & \footnotesize{LSTM-VAE} & \footnotesize{USAD} & \footnotesize{DAGMM} & \footnotesize{OmniAnomaly} & \footnotesize{AT} & \footnotesize{VT-SAT} & \footnotesize{VT-PAT} & \footnotesize{FATE} \\
\midrule

\multirow{3}{*}{PSM}  
& $\text{F1}_0$  & 51.42  & 52.36  & 50.16  & 50.89  & 50.17  & 29.18  & 50.15  & 49.68  & \textbf{\underline{65.72}}  \\
& $\text{F1}_1$  & 17.99  & 19.10  & 18.26  & 18.83  & 19.02  & 6.13   & 17.48  & 16.89  & \textbf{\underline{28.92}}  \\
& AUC            & 35.30  & 37.41  & 35.82  & 36.81  & 37.27  & 11.85  & 34.23  & 33.01  & \textbf{\underline{56.95}}  \\
\midrule

\multirow{3}{*}{SWaT}  
& $\text{F1}_0$  & 32.33  & 27.25  & 33.36  & 29.79  & 32.01  & 22.58  & 33.19  & 33.31  & \textbf{\underline{44.80}}  \\
& $\text{F1}_1$  & 15.06  & 12.88  & 13.51  & 13.19  & 13.03  & 1.02   & 15.06  & \textbf{\underline{15.26}}  & 13.47  \\
& AUC            & 30.02  & 25.66  & 26.91  & 26.34  & 26.38  & 1.39   & 30.09  & \textbf{\underline{30.50}}  & 27.27  \\
\midrule

\multirow{3}{*}{MSL}  
& $\text{F1}_0$  & 33.80  & 33.20  & 33.20  & 32.75  & 32.90  & 20.26  & 33.62  & 33.16  & \textbf{\underline{52.00}}  \\
& $\text{F1}_1$  & 8.03   & 8.05   & 8.24   & 8.06   & 8.21   & 1.69   & 8.50   & 8.53   & \textbf{\underline{21.54}}  \\
& AUC            & 14.16  & 14.21  & 14.71  & 14.47  & 14.63  & 2.29   & 15.17  & 15.12  & \textbf{\underline{42.47}}  \\
\midrule

\multirow{3}{*}{SMAP}  
& $\text{F1}_0$  & 24.79  & 24.58  & 25.81  & 28.37  & 28.09  & 19.54  & 22.92  & 22.94  & \textbf{\underline{41.74}}  \\
& $\text{F1}_1$  & 5.55   & 5.49   & 5.42   & 6.73   & 6.64   & 1.84   & 5.69   & 5.60   & \textbf{\underline{15.85}}  \\
& AUC            & 9.73   & 9.61   & 9.43   & 11.93  & 11.73  & 2.71   & 9.90   & 9.74   & \textbf{\underline{30.86}}  \\
\midrule

\multirow{3}{*}{SMD}  
& $\text{F1}_0$  & 50.87  & 50.61  & 50.48  & \textbf{\underline{51.00}}  & 50.90  & 10.69  & 48.84  & 48.15  & 33.86  \\
& $\text{F1}_1$  & 18.19  & 18.15  & 18.22  & \textbf{\underline{18.34}}  & 18.31  & 2.97   & 17.93  & 17.47  & 14.29  \\
& AUC            & 35.35  & 35.26  & 35.37  & \textbf{\underline{35.63}}  & 35.58  & 5.06   & 35.12  & 34.21  & 28.02  \\
\bottomrule
\end{tabular}%
}
\end{table*}

\begin{table*}[t]
\centering
\caption{Experimental results of the PA\%K metric for anomaly detection on five time-series datasets. The best scores are in bold and underlined.}
\label{tab:PA_K_results}
\renewcommand{\arraystretch}{0.4}
\setlength\tabcolsep{3pt}
\resizebox{\textwidth}{!}{%
\begin{tabular}{>{\centering\arraybackslash}p{1.2cm}|>{\centering\arraybackslash}p{1.2cm}|*{8}{>{\centering\arraybackslash}p{1.5cm}}|>{\centering\arraybackslash}p{1.5cm}}
\toprule
Dataset & Metric & \footnotesize{LSTM-AE} & \footnotesize{LSTM-VAE} & \footnotesize{USAD} & \footnotesize{DAGMM} & \footnotesize{OmniAnomaly} & \footnotesize{AT} & \footnotesize{VT-SAT} & \footnotesize{VT-PAT} & \footnotesize{FATE} \\
\midrule

\multirow{3}{*}{PSM}  
& $\text{F1}_{\text{PA}}$  & 92.94  & 94.51  & 92.22  & 92.33  & 92.24  & \textbf{\underline{97.78}}  & 92.68  & 91.84  & 96.01  \\
& F1                        & 35.96  & 37.88  & 34.98  & 35.91  & 35.30  & 2.31   & 30.26  & 33.80  & \textbf{\underline{46.79}}  \\
& AUC                       & 39.39  & 41.73  & 37.85  & 40.15  & 39.23  & 3.80   & 33.58  & 36.14  & \textbf{\underline{63.80}}  \\
\midrule

\multirow{3}{*}{SWaT}  
& $\text{F1}_{\text{PA}}$  & 86.26  & 86.46  & 87.28  & 85.55  & 87.33  & \textbf{\underline{96.65}}  & 84.65  & 84.63  & 88.99  \\
& F1                        & 77.13  & 76.57  & 76.57  & 76.57  & 76.57  & 3.45   & 78.00  & \textbf{\underline{78.19}}  & 74.73  \\
& AUC                       & 82.25  & 81.38  & 81.38  & 81.35  & 81.39  & 5.50   & 83.28  & \textbf{\underline{83.42}}  & 80.20  \\
\midrule

\multirow{3}{*}{MSL}  
& $\text{F1}_{\text{PA}}$  & 91.41  & 90.60  & \textbf{\underline{91.52}}  & 90.46  & 90.29  & 91.38  & 91.50  & 91.24  & 75.40  \\
& F1                        & 16.21  & 16.24  & 15.86  & 15.95  & 16.06  & 3.66   & 16.56  & 16.69  & \textbf{\underline{36.96}}  \\
& AUC                       & 24.80  & 24.90  & 24.44  & 24.83  & 24.87  & 5.61   & 25.92  & 25.72  & \textbf{\underline{51.08}}  \\
\midrule

\multirow{3}{*}{SMAP}  
& $\text{F1}_{\text{PA}}$  & 83.81  & 83.72  & 83.90  & 84.37  & \textbf{\underline{85.67}}  & 85.67  & 82.26  & 82.38  & 59.24  \\
& F1                        & 9.33   & 9.12   & 8.73   & 12.07  & 11.70  & 3.85   & 10.01  & 10.12  & \textbf{\underline{26.76}}  \\
& AUC                       & 14.48  & 14.21  & 14.34  & 18.78  & 18.25  & 6.08   & 16.25  & 16.50  & \textbf{\underline{37.07}}  \\
\midrule

\multirow{3}{*}{SMD}  
& $\text{F1}_{\text{PA}}$  & 86.61  & 86.61  & \textbf{\underline{86.66}}  & 86.61  & 86.63  & 77.68  & 86.41  & 86.36  & 61.31  \\
& F1                        & 46.35  & 46.24  & 47.09  & 46.69  & 46.66  & 2.66   & \textbf{\underline{48.46}}  & 47.30  & 30.03  \\
& AUC                       & 56.66  & 56.45  & 57.77  & 57.06  & 57.00  & 4.54   & \textbf{\underline{58.86}}  & 57.55  & 40.08  \\
\bottomrule
\end{tabular}%
}
\end{table*}

\autoref{tab:TaPR_results} and \autoref{tab:PA_K_results} report the anomaly detection performance of FATE and baseline methods using the TaPR and PA\%K metrics. These evaluations were conducted without incorporating the early-detection reward, thereby focusing purely on anomaly detection capability. FATE achieved strong performance on the PSM, MSL, and SMAP datasets, highlighting its effectiveness even in a fully unsupervised setting.

However, performance varied across datasets. On the SMD dataset, FATE was outperformed by several baselines, particularly on metrics that emphasize sustained detection over long anomaly intervals. This discrepancy can be attributed to the nature of the SMD dataset, which includes extended anomaly periods. While FATE effectively identifies the onset of anomalies via uncertainty, it may be less capable of persistently detecting long anomaly regions without access to reconstruction errors or target values.

Additionally, recent time-series forecasting models frequently incorporate normalization techniques such as RevIN \cite{RN46} to mitigate distribution shift. These methods can inadvertently obscure anomaly patterns by transforming anomalous regions to resemble normal behavior, further complicating uncertainty-based detection. Despite these challenges, FATE delivers competitive results compared to supervised or reconstruction-based approaches, even without using target labels. These results reaffirm FATE’s practical utility in real-world settings where labeled data is scarce and early warning is prioritized.

\subsection{Ablation Studies}
\label{sec:sec6.3}

\subsubsection{Sensitivity to Hyperparameters}
\label{sec:sec6.3.1}
\begin{figure}[!t]
    \centering  
    \includegraphics[width=\columnwidth]{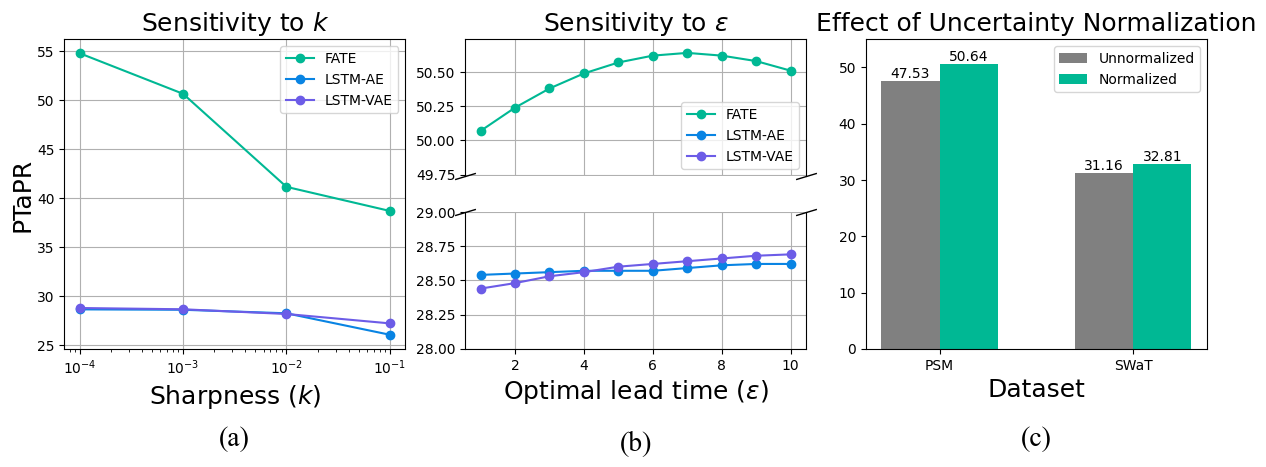}
    \caption{
Sensitivity analysis and ablation results of the FATE framework.  
(a) Sensitivity to the sharpness parameter $k$ on the PSM dataset.  
(b) Sensitivity to the optimal lead time $\epsilon$ on the PSM dataset.
(c) Effect of uncertainty normalization in FATE, showing consistent PTaPR improvement across both PSM and SWaT datasets.
}
    \label{fig:ablation}
\end{figure}
We conducted a sensitivity analysis of the PTaPR metric with respect to the sharpness parameter $k$ and the optimal lead time $\epsilon$ on the PSM dataset. As shown in \autoref{fig:ablation}~(a), decreasing $k$ gradually increases the PTaPR score (from $38.68$ at $k{=}0.1$ to $54.78$ at $k{=}0.0001$), as a smaller $k$ broadens the reward window for early predictions. To avoid over-rewarding excessively early signals, we set $k{=}0.001$ in the main experiments to strike a practical balance between responsiveness and realism.

In \autoref{fig:ablation}~(b), we observe that varying $\epsilon \in [1, 10]$ results in stable PTaPR scores, with peak performance at $\epsilon{=}7$. This suggests that the proposed framework is robust to moderate shifts in the assumed optimal lead time.

\subsubsection{Effect of Uncertainty Normalization}
\label{sec:sec6.3.2}
% \begin{figure}[!t]
%     \centering  
%     \includegraphics[width=\columnwidth]{Fig10_ablation.png}
%     \caption{Effect of uncertainty normalization in FATE on two datasets (PSM and SWaT). The PTaPR scores improved significantly after applying normalization, demonstrating the importance of aligning uncertainty scores across prediction horizons.}
%     \label{fig:Ablation}
% \end{figure}

We assess the impact of uncertainty normalization on PTaPR performance using the PSM and SWaT datasets. As shown in \autoref{fig:ablation}~(c), normalization significantly improves detection quality—raising PTaPR from 47.53 to 53.54 on PSM, and from 32.81 to 60.17 on SWaT.

These results indicate that normalization mitigates step-wise variance inflation, enables more consistent thresholding across prediction horizons, and enhances overall precursor detection effectiveness.

\subsection{Visualization}
\label{sec:sec6.4}

\subsubsection{Visualization of Anomaly Detection}
\label{sec:sec6.4.1}
\begin{figure*}[!t]
    \centering  
    \includegraphics[width=\textwidth]{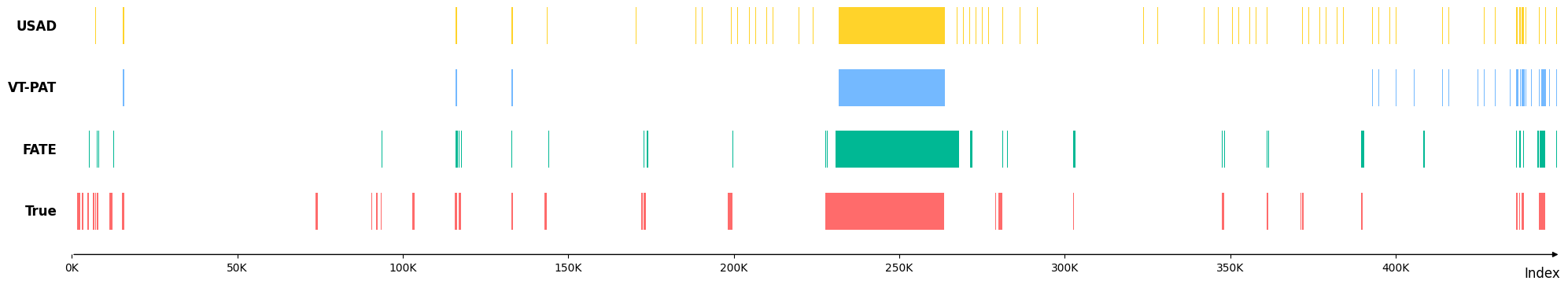}
    \caption{Visualization of anomaly detection results by different models on the SWaT dataset. Ground-truth anomalies are shown in red, with detected segments overlaid for comparison.}
    \label{fig:visualization_detected}
\end{figure*}

\autoref{fig:visualization_detected} compares the anomaly segments detected by the proposed model FATE and two baseline methods on the SWaT dataset: USAD, which performs well under the PTaPR metric, and VT-PAT, which ranks highly under the PA\%K metric. The bottom panel shows the ground-truth anomaly regions (red), while each model's detections are color-coded across the timeline.

USAD exhibits frequent over-detection, especially in the latter half of the sequence, resulting in many false positives. VT-PAT shows a more stable detection pattern but fails to capture several key anomaly intervals. In contrast, FATE closely aligns with the actual anomaly segments and significantly reduces false alarms. For example, in the anomaly event occurring after index 250,000, FATE captures the region with high accuracy, demonstrating precise temporal localization.

These results highlight FATE's ability to detect various anomaly patterns robustly. Its reduced false-positive rate and strong agreement with ground truth suggest its practical utility in real-time early warning systems based on time-series data.

\subsubsection{Visualization of Precursor-of-Anomaly Detection}
\label{sec:sec6.4.2}
\begin{figure}[t!]
    \centering  
    \includegraphics[width=\columnwidth]{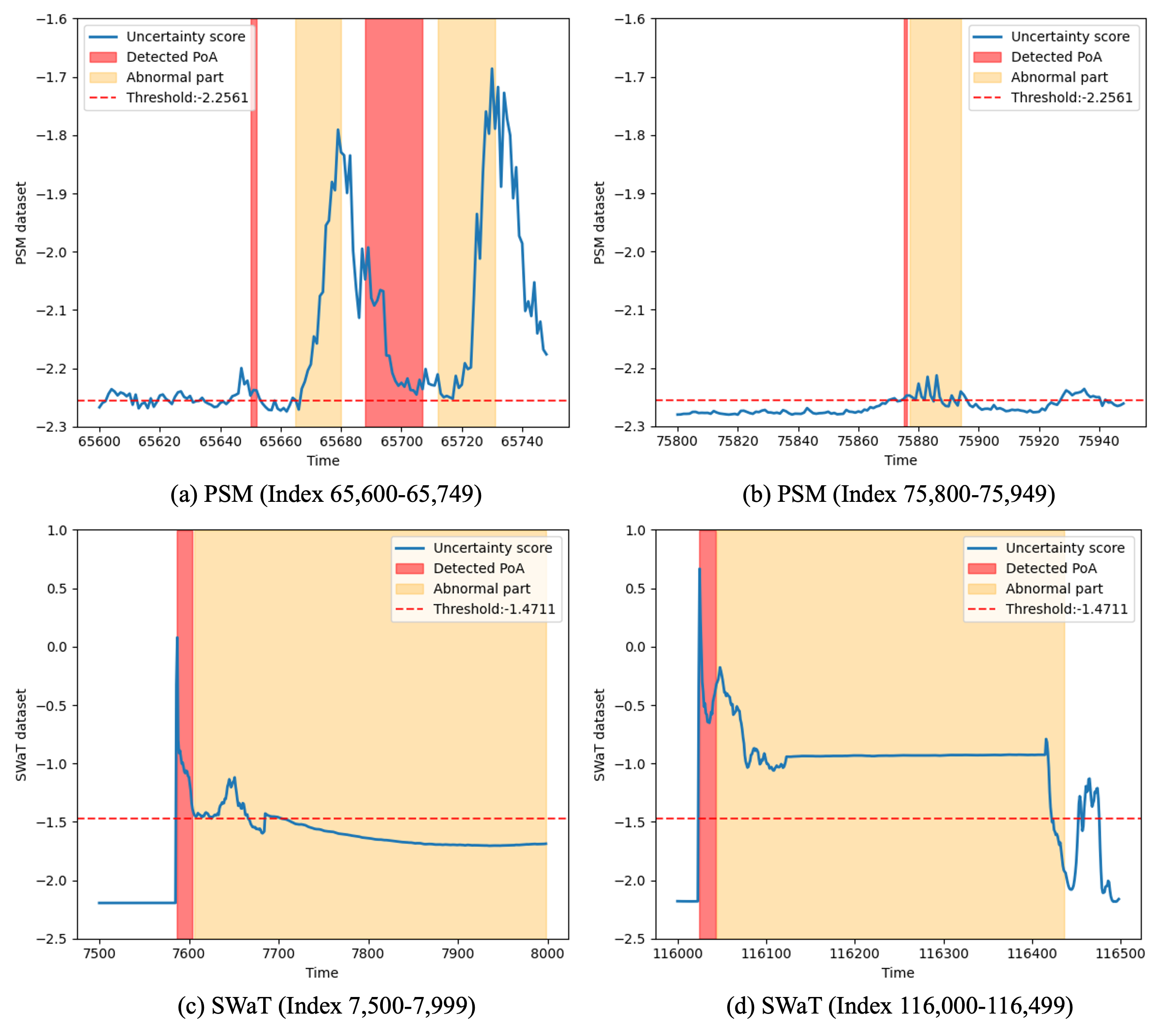}
    \caption{Visualization of Precursors-of-Anomaly (PoA) detected by FATE on the PSM and SWaT datasets. The blue line shows the uncertainty scores, red regions indicate detected PoA segments, and yellow regions mark ground-truth anomalies.}
    \label{fig:visualization_PoA}
\end{figure}

\autoref{fig:visualization_PoA} illustrates the ability of FATE to detect PoA prior to their onset on the PSM and SWaT datasets. The figure contains four subplots, each highlighting a specific time interval. Subplots (a) and (b) show results for PSM, while (c) and (d) present those for SWaT.

In each plot, the blue line denotes the model’s uncertainty score, the red-shaded areas indicate detected PoA segments, and the yellow bars mark ground-truth anomalies. A red dashed line represents the detection threshold. On the PSM dataset, the uncertainty score increases markedly before or at the early stage of anomaly onset, successfully triggering precursor detection ahead of time. On the more complex SWaT dataset, the model similarly captures early warning signals: the uncertainty rises just before the anomalies begin, and the detected PoA segments exhibit strong temporal alignment with the true anomaly intervals.

These results demonstrate FATE's effectiveness in leveraging uncertainty for early anomaly prediction. The model not only anticipates anomalies with high temporal precision but also minimizes false alarms. This capability suggests strong applicability to real-time monitoring systems requiring proactive failure prevention.

\section{Conclusion}
\label{sec:sec7}

Time-series anomaly detection is crucial for ensuring system reliability and preventing unexpected failures. While most existing methods detect anomalies only after their occurrence, such retrospective detection limits proactive responses. In contrast, precursor-of-anomaly (PoA) detection seeks to identify early warning signs before anomalies occur, though it remains a challenging task due to the absence of future observations.

To this end, we propose FATE, an uncertainty-aware ensemble forecasting framework that identifies anomaly precursors based on the variance of predictions from multiple models. FATE enables early anomaly detection without relying on future target values, operating fully in an unsupervised setting. Additionally, we introduce PTaPR, a novel evaluation metric that extends TaPR to more accurately assess timeliness and precision in precursor detection. Experiments on multiple datasets demonstrate that FATE consistently outperforms baselines in both standard and precursor-oriented metrics.

While effective, FATE's ensemble-based design introduces computational overhead due to multiple model training and inference. To mitigate this, future work will investigate single-model approaches to uncertainty estimation, such as Bayesian neural networks, which can enhance the scalability and practicality of PoA detection in real-time applications.


\bibliographystyle{IEEEtran}
\bibliography{fate_references}

% \newpage

 

\begin{IEEEbiography}[{\includegraphics[width=1in,height=1.25in,clip,keepaspectratio]{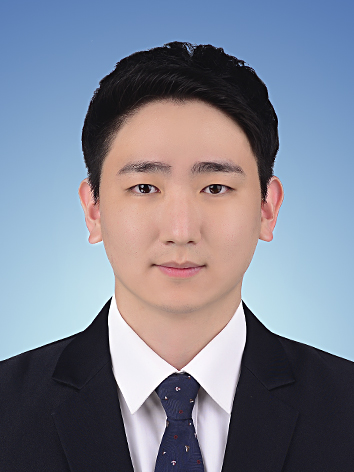}}]{Hyeongwon Kang} is currently pursuing a Ph.D. degree in Industrial and Management Engineering at Korea University, Republic of Korea. He received a B.S. degree in Industrial and Management Engineering and Artificial Intelligence (double major) from Incheon National University, Republic of Korea. His research focuses on time-series machine learning, with particular interests in anomaly detection, forecasting, and interpretable AI for industrial applications.
\end{IEEEbiography}

\begin{IEEEbiography}[{\includegraphics[width=1in,height=1.25in,clip,keepaspectratio]{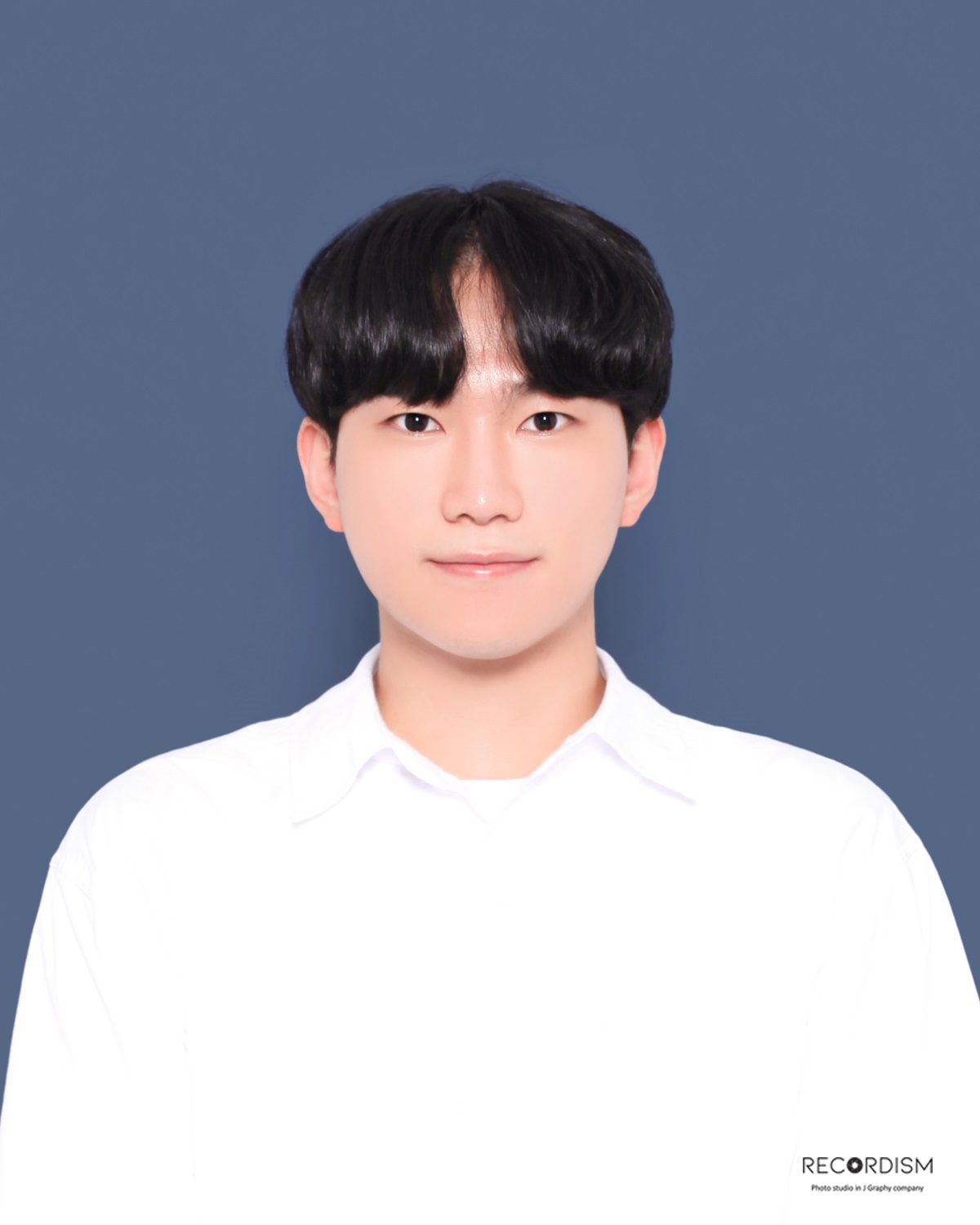}}]{Jinwoo Park} is currently pursuing a Ph.D. degree in the Department of Industrial Engineering at Seoul National University, Republic of Korea. He received an M.S. degree in Industrial and Management Engineering from Korea University and a B.S. degree in Financial Engineering from Ajou University, Republic of Korea. His research interest is developing machine learning algorithms for time-series analysis, focusing on prediction and anomaly detection.
\end{IEEEbiography}

\begin{IEEEbiography}[{\includegraphics[width=1in,height=1.25in,clip,keepaspectratio]{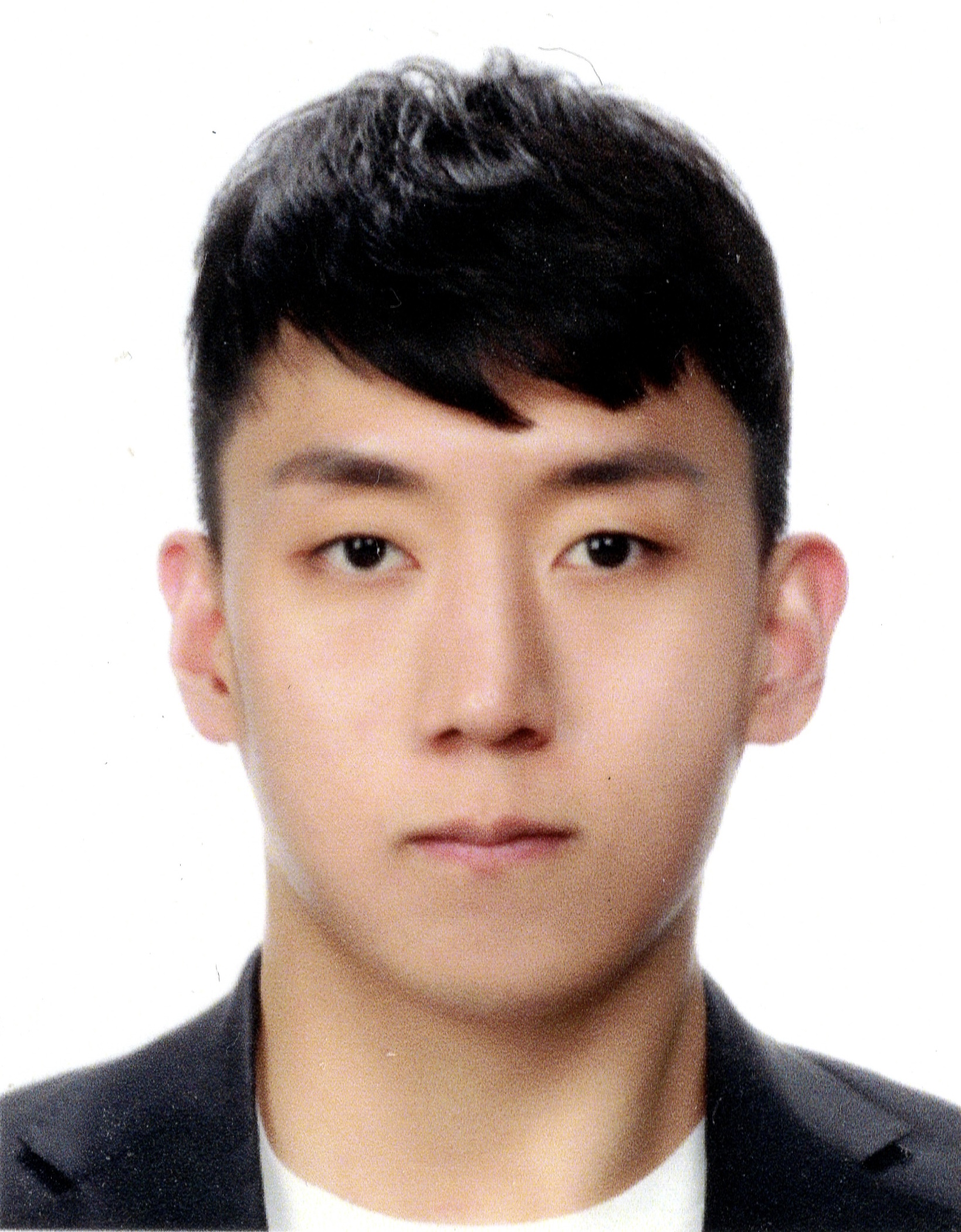}}]{Seunghun Han} is currently employed as an AI scientist by LG CNS. He received a B.S. in Statistics and an M.S. in Industrial and Management Engineering from Korea University, Republic of Korea. His research interest is in developing machine learning algorithms for time-series analysis, focusing on prediction and anomaly detection.
\end{IEEEbiography}

\begin{IEEEbiography}[{\includegraphics[width=1in,height=1.25in,clip,keepaspectratio]{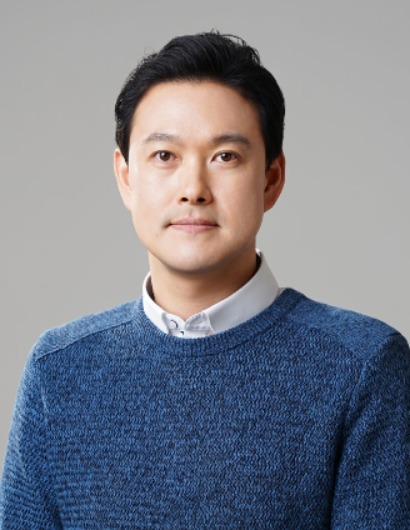}}]{Pilsung Kang} is an Associate Professor in the Department of Industrial Engineering at Seoul National University, South Korea. He received his B.S. and Ph.D. in Industrial Engineering from Seoul National University. His research focuses on developing machine learning algorithms for structured and unstructured data (including image, video, and text), with applications in areas such as fault classification in manufacturing, anomaly detection from system logs, and sentiment analysis of news and review texts.
\end{IEEEbiography}

% \section*{Acknowledgments}
% This should be a simple paragraph before the References to thank those individuals and institutions who have supported your work on this article.

\end{document}